\documentclass[sn-mathphys-num]{sn-jnl}

\usepackage{graphicx}%
\usepackage{multirow}%
\usepackage{amsmath,amssymb,amsfonts}%
\usepackage{amsthm}%
\usepackage{mathrsfs}%
\usepackage[title]{appendix}%
\usepackage{xcolor}%
\usepackage{textcomp}%
\usepackage{manyfoot}%
\usepackage{booktabs}%
\usepackage{algorithm}%
\usepackage{algorithmicx}%
\usepackage{algpseudocode}%
\usepackage{listings}%
\usepackage{todonotes}%
\usepackage{adjustbox}
\usepackage{comment} 

\theoremstyle{thmstyleone}%
\theoremstyle{thmstyletwo}%

\theoremstyle{thmstylethree}%
\newtheorem{definition}{Definition}%

\makeatletter
\renewcommand{\fnum@algorithm}{\fname@algorithm~\thealgorithm}
\renewcommand{\fname@algorithm}{}
\makeatother

\begin{document}

\title[Effective Non-Random Extreme Learning Machine]{Effective Non-Random Extreme Learning Machine}


\author[1]{\fnm{Daniela} \sur{De Canditiis}}\email{d.decanditiis@iac.cnr.it}
\equalcont{These authors contributed equally to this work.}

\author*[2]{\fnm{Fabiano} \sur{Veglianti}}\email{fabiano.veglianti@uniroma1.it}
\equalcont{These authors contributed equally to this work.}


\affil[1]{\orgdiv{Istituto per le Applicazioni del Calcolo}, \orgname{CNR}, \orgaddress{\street{via dei Taurini 19}, \city{Rome}, \postcode{00185}, \country{Italy}}}

\affil[2]{\orgdiv{Department of Computer, Control, and Management Engineering Antonio Ruberti}, \orgname{Sapienza University of Rome}, \country{Italy}}



\abstract{The Extreme Learning Machine (ELM) is a growing statistical technique widely applied to regression problems. In essence, ELMs are single-layer neural networks where the hidden layer weights are randomly sampled from a specific distribution, while the output layer weights are learned from the data. Two of the key challenges with this approach are the architecture design, specifically determining the optimal number of neurons in the hidden layer, and the method’s sensitivity to the random initialization of hidden layer weights.

This paper introduces a new and enhanced learning algorithm for regression tasks, the Effective Non-Random ELM (ENR-ELM), which simplifies the architecture design and eliminates the need for random hidden layer weight selection. The proposed method incorporates concepts from signal processing, such as basis functions and projections, into the ELM framework. We introduce two versions of the ENR-ELM: the approximated ENR-ELM and the incremental ENR-ELM. Experimental results on both synthetic and real datasets demonstrate that our method overcomes the problems of traditional ELM while maintaining comparable predictive performance.}




\maketitle

\begingroup\small
\noindent%
This preprint has not undergone peer review or any post‑submission improvements or corrections. 
The Version of Record of this article is published in Neural Computing and Applications, and is available online at \url{https://doi.org/10.1007/s00521-025-11519-5}
\par\endgroup

\section{Introduction}\label{sec1}
ELM is an emerging statistical technique widely used for regression, classification, clustering,  sparse approximation, feature learning, and compression.  In this article, we focus on regression to simplify the discussion, but the idea presented can be adapted to other learning tasks. The most notable advantage of ELM is that there is no requirement for the hidden layer's weights to be tuned. The hidden layer is randomly generated and is never updated after that. ELMs have been largely explored in the last decade; the reader can refer to \citep{scardapane2017} and \cite{survey2023IEEE} for interesting overviews. 
ELMs have been applied in many applicative contexts, but here we are more interested in the methodological point of view, therefore we do not assume any special conditions on the data. Furthermore, we would like to emphasize that the performance of an ELM compared to a traditional neural network with the same architecture has already been analyzed in the interesting work \citep{NCA2021}. In \citep{NCA2021} the authors show that there are cases in which ELMs have better accuracy than traditional neural networks, but above all ELMs are more convenient in terms of computation time, as they do not require the iterative back-propagation algorithm to learn weights. Our research aims to propose an alternative methodology for applying ELMs when they are preferred over other estimators, without claiming to propose a universal method that is suitable for every case. In this regard, we cite the same paper \citep{NCA2021} ``[Experimental results] confirm that for relatively small datasets, ELMs are better than network trained by the backpropagation algorithm", meaning that ELMs are preferred when the dimension of the input space is relatively small, hence this will be our working assumption.


\subsection{Motivation}

The motivation for this paper stems from both our practical experience and a thorough review of the current literature. Providing a comprehensive summary of the existing research is challenging because many articles discuss ELMs and potential improvements. However, we believe that two recent papers [1] and [4] offer solid overviews of both the strengths and limitations of this approach, highlighting areas that require further attention. Based on our own studies, we have identified three key issues in the application of ELMs, which form the foundation of our research rationale. 

\noindent 
\textbf{Motivation (i)}  From the recent review \citep{scardapane2017} we cite ``[ELM] still suffer from design choices, translated in free parameters,
which are difficult to set optimally with the current mathematical framework, so practically
they involve many trials and cross-validation to find a good projection space [number of hidden neurons]." The first motivation for our research is then the lack of an effective model selection criterion, or in other words, the lack of an effective architecture design.

\noindent 
\textbf{Motivation (ii)} In the same review \citep{scardapane2017}, we cite ``[ELM] ... on top of the selection of the number of hidden [neurons] and the nonlinear [activation] function strongly depend on the particular random projection".  In \citep{IEEE2017}, we cite ``The supervisory mechanism reported in this paper clearly reveals that the assignment of random parameters of learner models should be data dependent to ensure the universal approximation property of the resulting randomized learner model." One of the more interesting attempts at a solution to this issue is given by \citep{Neurocomputing2011}, where a data-dependent weights matrix for the hidden layer is proposed. Specifically in \citep{Neurocomputing2011} a proper weights matrix that sorts the input data $\bold x_i$ by a well defined \emph{lexicographical} order is established, the resulting hidden layer output matrix is proved to be non-singular, then, for the evaluation of the output weights, a classical OLS algorithm is applied. However, the proposal in \citep{Neurocomputing2011} is specific for radial basis function activation and does not include model selection. The second motivation for our research is then to provide a data-dependent deterministic mechanism to fix the hidden layer output matrix, justifying the name ``Non-Random" for our proposed method.

\noindent 
\textbf{Motivation (iii)} In \citep{NCA2021}, we cite ``There are no smart algorithms for the inverse matrix calculation so that determining weights in the output layer for challenging dataset become feasible and memory efficient." The third motivation for our research is then to furnish a very smart and efficient algorithm for determining weights in the output layer without employing matrix inversion. 

\subsection{Contribution}
In this paper, we propose the Effective Non-Random ELM (ENR-ELM), a new method for non-parametric regression tasks. This method integrates signal processing concepts, such as basis functions and projections, into the ELM framework. The ENR-ELM algorithm operates in two phases: (1) the selection of a data-dependent hidden layer's weights matrix and (2) the evaluation of the output layer weights.

In phase (1), the expected Gram matrix of the input data vectors $\bold x_i$, transformed by the hidden layer's random feature map, is computed along with its eigen-decomposition. The orthonormal matrix $U$, whose columns are the eigenvectors of the Gram matrix, is then used to derive an optimal weights matrix $\hat{W}$, which represents the data-dependent selection of hidden layer's weights.

Phase (2) can be implemented in two ways, resulting in two algorithms: the approximate ENR-ELM and the incremental ENR-ELM. In the first algorithm, the matrix $U$ is used as a proxy for the hidden layer output matrix, while in the second, the hidden layer output matrix is used in a classical incremental forward-stagewise regression. In both cases, the test error evaluation guides the selection of the optimal number of columns, corresponding to the appropriate number of neurons in the hidden layer.

To determine the optimal weights matrix $\hat{W}$ in phase (1) of the procedure, the network's activation function must be invertible and take values in the range $[-1, 1]$. Therefore, in the examples presented in this paper, we use the \emph{Erf} activation function. The key distinction between our ENR-ELM and the traditional ELM lies in the choice of the hidden layer weight matrix, which is data-dependent rather than random in our approach. Although the spectral decomposition of the Gram matrix is computationally intensive, this cost is offset by the linear complexity of the second phase, making the overall procedure more efficient compared to the traditional ELM, particularly in terms of model selection.

\subsection{Paper structure}

The remainder of the paper is organized as follows: In Section \ref{sec:background}, we review the Extreme Learning Machine (ELM) and the Neural Network Gaussian Process (NNGP) kernel, which arises when considering random neural networks in the infinite-width regime. Section  \ref{sec: mehtod} introduces the proposed ENR-ELM method and outlines its pseudo-algorithm. Extensive experimental results are presented in Section \ref{sec: experimental}. Finally, Section \ref{sec: conclusion} summarizes our conclusions and discusses future research directions.

\section{Background}\label{sec:background}

\subsection{Notations}
We denote the identity matrix of dimension $d$ by $I_d$. We use bold notation to denote a vector, which should be considered as a column vector unless stated otherwise. We denote matrices by capital letter and the $j$-th column of matrix $H$ by $H_{\cdot j}$, the $i$-th row of matrix $H$ by $H_{i \cdot}$; moreover if $J$ is a set of indices, then $H_{\cdot J}$ represents the submatrix of $H$ restricted to the columns with indices in $J$.
The symbol $J^c$ denotes the complement of the set $J$. Finally, the following notation applies $\bold{1}_n=(1,\ldots,1) \in \mathbb{R}^{n} $.

\subsection{ELM}
Given a vector $ \bold x \in \mathbb R^{n_0}$, its random projection is defined as
\begin{equation} \label{eq: random_proj}
z(\bold x)=(z_1(\bold x),\ldots,z_n(\bold x)) :=\sigma( W \, \bold x )  \in \mathbb R^{n}
\end{equation}

\noindent 
where $W \in \mathbb R^{n \times n_0}$ is a random matrix with independent entries $W_{ij} \sim {\cal{N}}(0,1/n_0)$, and $\sigma:\mathbb{R}\to\mathbb{R}$ is referred to as the activation function, applied entry-wise to the vector $W\,\bold x$. Given the activation function $\sigma$, the weights matrix $W$, and a weights vector $\boldsymbol{\beta} \in \mathbb{R}^{n}$, a shallow fully connected neural network is the following function 
\[
\bold x:=(x_1,\ldots,x_{n_0})\in\mathbb R^{n_0}\mapsto f_{(\sigma, W, \boldsymbol{\beta})}(\bold x):= \boldsymbol{\beta}^\top \sigma( W \,\bold x ) \,  \in \mathbb{R}.
\]

\noindent 
Consider a regression task with training sample $\{ ( \bold x_i, y_i ) \}_{i=1,...,T}$, where $\bold x_i \in \mathbb R^{n_0}$ and $y_i \in \mathbb R$. The ELM first maps the input vector $\bold x$ into a high (theoretically even infinite) dimensional feature space, then applies a linear model into this feature space producing the following non-linear estimator 
\begin{equation} \label{eq: regression_function}
\hat{f}_{(\sigma, W, \hat{\boldsymbol{\beta}})}(\bold x) = \sum_{i=1}^n \hat{\beta}_i z_i(\bold x) = \hat{\boldsymbol{\beta}}^\top    \,\sigma(W \, \bold x ) ,
\end{equation}

with $\hat{\boldsymbol{\beta}}$ learned from data by minimizing the following empirical risk
\begin{equation} \label{eq: betahat_def}
\hat{\boldsymbol{\beta}}= \text{argmin}_{\boldsymbol{\beta} \in \mathbb{R}^{n}} \, \, \sum_{l=1}^T \left(y_l -  \boldsymbol{\beta}^\top \sigma(  W\, \bold x_l) \right)^2.
\end{equation}

To clarify, ELM is a 1-layer (shallow) neural network where only the weights of the output layer are learned from the data, while the weights in the hidden layer are sampled from a Gaussian distribution and not tuned. 

\vspace{0.5cm}

\noindent

Consider data matrix $X=[\bold x_1 , \bold x_2 , \ldots ,\bold x_T] \in \mathbb{R}^{ n_0 \times T}$, hidden layer output matrix $S=\sigma(WX)^\top=[\sigma( W \, \bold x_1)^\top, \,  \,\sigma(W \, \bold x_2)^\top, \,  \ldots \,, \sigma(W \, \bold x_T)^\top] \in \mathbb{R}^{T \times n}$ and data vector $\bold y=[y_1,y_2,...,y_T] \in  \mathbb{R}^{T}$, then in matrix notation problem in eq.~\eqref{eq: betahat_def} becomes 
\begin{equation} \label{eq: betahat_defmatrix}
\hat{\boldsymbol{\beta}}= \text{argmin}_{\boldsymbol{\beta} \in \mathbb{R}^{n}} \, \, \left\| \bold y - S \boldsymbol{\beta} \right\|_2^2
\end{equation}
a linear model with stochastic design matrix $S$. 

If the matrix $S^\top \, S $ has full rank, then 
the solution is given by
\begin{equation} \label{eq: lowdim}
\hat{\boldsymbol{\beta}}=(S^\top \, S )^{-1} S^\top y,
\end{equation}
otherwise, the solution can be obtained by different approaches. It is possible to use the incremental forward-stagewise regression or penalized regression approaches such as Ridge regression or LASSO regression. In the case of Ridge regression the minimization given in eq.~\eqref{eq: betahat_defmatrix} is penalized by the $l_2$ norm of the regression coefficients and the solution is given by 
\begin{equation} \label{eq: highdim}
\hat{\boldsymbol{\beta}}=(S^\top \, S +\lambda I_n)^{-1} S^\top y = S^\top ( S \, S^\top+\lambda I_T)^{-1} y,
\end{equation}
where the positive parameter $\lambda$ is chosen to avoid overfitting and preserve a good generalization property. Sometimes this choice is replaced by an early stopping criterion of the gradient descent method applied to minimization in eq.~\eqref{eq: betahat_defmatrix}. While the incremental forward-stagewise regression or the LASSO regression guarantees variable selection (which is important for determining the relevants hidden layer neurons), the Ridge regression does not guarantee variable selection. In addition, we note that evaluating the expression in eq.~\eqref{eq: lowdim} results in a computational cost that is cubic in the number of features, while evaluating the second expression in eq.~\eqref{eq: highdim}  results in a computational cost that is cubic in the number of data points, which can be unfeasible in a data rich situation.

The choice of the number of features $n$ (i.e. the number of neurons in the hidden layer) is important and significantly depends on the structure of the input dataset $X$. As explained in the introduction, we do not consider any specific type of data, but our working hypothesis is that the input space dimension $n_0$ is relatively small. 
Therefore, we can think the input space to be an Euclidean space of limited dimension, and the random feature map (defined in eq.~\eqref{eq: random_proj}) as a mechanism for scattering the data into a much higher-dimensional space, enhancing certain data characteristics. It is somewhat like the opposite of dimensionality reduction; here, the data dimensionality is increased to reveal hidden structures, allowing a linear operator to perform effectively in the new space. Mathematically, the data structure is captured by the Gram matrix. A significant area of research is understanding how the Gram matrix of the mapped features, $S S^\top \in \mathbb{R}^{T \times T}$, relates to the Gram matrix of the input features, $X^\top X \in \mathbb{R}^{T \times T}$. Some results in this area, specific to the case of ReLU activation function, are provided in  \citep{IEEE2016}.
The choice of the number $n$ of hidden neurons, which in this paper is referred to as model selection or architecture design, is a critical aspect; practitioners usually select $n$ by using the test error or some cross-validation criterion. One effective attempt to solve this problem is the incremental constructive approach proposed in \citep{2017IEEE}, where a starting small-sized network is incrementally enlarged by generating hidden layer neurons and output weights until a predefined termination criterion is met.

\noindent As pointed out in motivation (i), there is still a lack of effective model selection criteria for ELM; our proposed method address this issue.

\noindent Moreover, as described in motivation (ii), beyond the choice of the number of hidden neurons $n$, ELM performance is conditioned by the particular sampling of the random matrix $W$; hence, some machine learning practitioners generate independent realizations of the ELM and subsequently apply some voting strategy, for example in \citep{voting2012IS}. 
Our proposed method accounts for this issue too, since the weights matrix $W$ will be learned from data and will not be stochastic.

\noindent Finally, as described in motivation (iii), matrix inversion in eq. ~\eqref{eq: lowdim} or eq.~\eqref{eq: highdim} is computationally expensive if one considers that for each choice of $n$ and for each realization of $W$ it must be performed from scratch. Our proposed method also alleviates this issue because, apart from the eigen-decomposition of the Gram matrix, it does not require any matrix inversion to evaluate the coefficients in the last layer for different choices of $n$.

\subsection{Neural Network Gaussian Process Kernel}

In recent years, there has been a significant body of research focused on neural networks with stochastic weights in the infinite-width regime. Since the seminal work published in 1996 \citep{Neal1996}, multiple studies have shown that a neural network with independent Gaussian weights converges to a Gaussian process with zero mean and a known covariance matrix as the number of neurons in the hidden layers approaches infinity. While the theoretical understanding of this topic is quite extensive, see e.g. \citep{Hanin2023}, \citep{Marinucci2023}, \citep{Apollonio2024},
the practical knowledge for using these results is still in its infancy. In this regard, a software package capable of calculating the limiting process covariance matrix, called Neural Network Gaussian Process (NNGP) kernel, has been made available to the scientific community for the first time in the paper \citep{ingegnerigoogle2024}.
For completeness, in the following we provide the definition of NNGP kernel given in \citep{ingegnerigoogle2024} for a general deep neural network.

Consider a fully connected L-layered neural network $f : \mathbb{R}^{n_0} \rightarrow \mathbb{R}$ for $L \geq 1$ defined as 
\begin{equation} \label{eq: FCNN}
f_{(\sigma, \mathcal{W})}(\bold x)= {\bold w^{(L+1)}}^\top z^{(L)}(\bold x) , \quad z^{(l)}(\bold x)=\sigma(W^{(l)} z^{(l-1)}(\bold x) ), \quad z^{(0)}(\bold x)=\bold x,
\end{equation}
 where $\mathcal{W}:= vec(\bold w^{(L+1)}, \cup_{l=1}^L W^{(l)})$, with $\bold w^{(L+1)} \in \mathbb{R}^{n_L}$ and $W^{(l)} \in \mathbb{R}^{n_{l} \times n_{l-1}}$ for $l \geq 1$, is the collection of all weight parameters. Note that the definition given in eq.~\eqref{eq: FCNN} coincides with the definition given in eq.~\eqref{eq: regression_function} when $L=1$ and $\mathcal{W}:= vec(\boldsymbol{\beta}, W)$, being the weights of the output layer $\bold w^{(L+1)}= \boldsymbol{\beta}$ and weights of the hidden layer $W^{(1)}=W$. 
 
\noindent In \citep{ingegnerigoogle2024}, for simplicity, a fully connected L-layered neural network with the same number of neurons in the hidden layer, i.e. $n_{l}=n$ for all $l=1,..., L+1$ is considered. This assumption is not necessary to achieve the convergence results presented in the literature which hold for $\text{min} \{n_1,...,n_{L+1}\} \rightarrow \infty$, see \citep{Hanin2023}, however, this hypothesis simplifies exposition.

\vspace{0.5cm}

\begin{definition}[Definition NNGP Kernel]
Consider a stochastic fully connected L-layered neural network, as the one given in eq.~\eqref{eq: FCNN} where all elements of $\mathcal{W}$ are independent Gaussian variables with zero mean and variance $1/n$, except in the first hidden layer where the variance is $1/n_0$, then the NNGP kernel is the covariance of the limiting Gaussian process defined as
\begin{equation}
\Sigma_{\sigma}^{L+1}(\bold x,\bold y):= plim_{n \rightarrow \infty} < f_{(\sigma, \mathcal{W})}(\bold x) , f_{(\sigma, \mathcal{W})}(\bold y) >, \quad \quad \forall \, \bold x, \bold y \in \mathbb{R}^{n_0},
\end{equation}
where $plim$ stands for convergence in probability. 
\end{definition}

\vspace{0.3cm}

\noindent Note that in the original paper \citep{ingegnerigoogle2024}, the weights are all independent and identically distributed $\mathcal{N}(0,1)$ because in the definition of a fully-connected L-layered neural network given in eq.~\eqref{eq: FCNN} the matrix of weights at layer $l$ is rescaled by the factor $1/\sqrt{n_{l-1}}$.

\noindent In \citep{ingegnerigoogle2024} the authors proposed a fast algorithm that evaluates (for activation functions admitting exact dual activation, like erf,ReLU, etc) or approximates (for those activation functions not admitting exact dual activation) any L-layered NNGP kernel matrix. This algorithm is available at \url{https://github.com/google/neural-tangents} as a Python package; its manual furnish the complete list of activation functions for which to evaluate or approximate the NNGP kernel. The method described in \citep{ingegnerigoogle2024} is based on the following  recursive formula for the NNGP kernel: start from the input layer Kernel $K_{\sigma}^{(0)}(\bold x,\bold y):=<\bold x,\bold y>$ and define the successive kernels, for $l=1,..., L+1$, by 
\[
K_{\sigma}^{(l)}(\bold x,\bold y):= \mathbb{E}_{(u,v) \sim \mathcal{N}(0, \Lambda_{\sigma}^{(l)})} (\sigma(u) \sigma(v))
\]
where the covariance matrix is $ \Lambda_{\sigma}^{(l)}:= \left[ \begin{array}{cc}
K_{\sigma}^{(l-1)}(\bold x,\bold x) & K_{\sigma}^{(l-1)}(\bold x,\bold y) \\
K_{\sigma}^{(l)}(\bold x,\bold y) & K_{\sigma}^{(l)}(\bold y,\bold y)
\end{array}  \right] \in \mathbb{R}^{2} $.

From our point of view, this package will be used to evaluate the NNPG kernel in the input data $\{ \bold x_i \}_{i=1,...,T}$, which represents the Gram matrix of their transformation produced by the hidden layer of a stochastic shallow neural network with an infinite number of neurons.

\section{The proposed method}\label{sec: mehtod}
Let us recall the well-known correspondence between deep neural networks and kernel methods, particularly the relationship between Extreme Learning Machines (ELMs) and kernel methods, as described in detail in \citep{IEEEPR2019}.

\noindent
In kernel methods, there is no need to explicitly define the feature map because the kernel, a non-negative definite bivariate function, is used to compute the Gram matrix of the transformed sample data and to generalize the estimator to new data. This ability to rely solely on the kernel, bypassing the explicit computation of the feature map, is referred to as the ``kernel trick" in machine learning.
On the other hand, in ELMs, the feature map — defined in eq.~\eqref{eq: random_proj}, which projects the input data $\bold x \in \mathbb{R}^{n_0}$ into a high-dimensional (of order $n$) or even infinite-dimensional space ($n \rightarrow \infty$) — is explicitly constructed, avoiding the use of the kernel. In both methods, the estimator is a linear combination of the features that have been transformed using the map and only the coefficients of this linear combination need to be learned from the data.
The method proposed in this research combines these two approaches. As in kernel methods, the kernel is fixed, arguably the optimal choice you can have when using random feature map given in  eq.~\eqref{eq: random_proj}; then we explicitly construct the optimal feature map, which determines the hidden layer output matrix of a shallow neural network. Subsequently, only the weights of the final layer need to be learned, as in ELMs.

We now have all the ingredients to present our proposed method. Consider a random shallow neural network with an infinite number of neurons in its hidden layer, and consider the NNGP kernel $K_{\sigma}=(K_{\sigma}^{(1)}(\bold \bold x_i,\bold \bold x_j))_{i,j=1,...,T} \in \mathbb{R}^{T \times T}$ evaluated at the training input vectors by the exact method presented in Section \ref{sec:background}. Matrix $K_{\sigma}$ represents the expected Gram matrix of the input data when transformed through the hidden layer of a random shallow neural network with an infinite number of neurons. Matrix $K_{\sigma}$ is known to be non-negative defined, let $K_{\sigma}=U \, \Delta \, U^\top$ be its spectral decomposition and rearrange the columns of matrix $U$ according to their informative content, that is, ordering them based on the quantity  $|  < y, U_{.j}>|$ arranged in decreasing order.

In this way, for any $n \leq T$, the first $n$ columns of the eigen-matrix $U$ represent an orthonormal basis of the best (most informative) subspace of dimension $n$ into which the input data is mapped by the hidden layer transformation when using $n$ neurons.
Although we can immediately evaluate the training vector prediction, as $\hat{\bold y}=U\, U^\top \bold y$, to generalize the proposed estimator we need an oracle providing us the feature map $\sigma(W \bold x)$ such that $\sigma( W \bold x_i)=U_{i.}$. The oracle does not exist, but we can find an optimal matrix $\hat{W} \in \mathbb{R}^{n \times n_0}$  such that $\sigma(\hat{W} X)^\top $ is similar to matrix $ U $ as much as possible. We propose to solve the following
\begin{equation} \label{eq: defW}
\hat{W} =argmin_{W \in \mathbb{R}^{T \times n_0}}  \left\| \sigma(W X)^\top - U   \right\|^2. 
\end{equation}

\noindent
 Since the columns of $U$ are by construction orthonormal, their entries are within the range $[-1,1]$; assuming that the activation function is invertible and takes values in $[-1,1]$, we obtain the solution of equation \eqref{eq: defW} in a closed form, i.e. $\hat{W}=\sigma^{-1}(U^\top) \, X^\top \, (X X^\top)^{-1}$, where $\sigma^{-1}$ is intended to be applied entry-wise. After evaluating the optimal matrix $\hat{W}$ we have two approaches. In the first approach, which we call A-ENR-ELM (approximated ENR-ELM), we learn the output layer coefficients using the OLS to regress $y$ on the matrix $U$, which is a proxy for the matrix $\sigma(\hat{W} X)^\top$. This approach computes the final layer coefficients $\hat{\boldsymbol{\beta}}=U^\top y$ with linear computational cost. In the second approach, we learn the output layer coefficients using the classical incremental forward stagewise regression of $y$ on matrix $S=\sigma(\hat{W} X)^\top$, for that reason we call this second approach I-ENR-ELM (incremental ENR-ELM). In this second approach, each neuron (which corresponds to a column of matrix $S$) is partially added to the estimator incrementally until some convergence criterion is achieved.

 \noindent
 Finally, knowing matrix $\hat{W}$ and vector $\hat{\boldsymbol{\beta}}$, obtained by one of the two approaches, it is possible to generalize the proposed estimator; indeed given a new input data $\bold x \in \mathbb{R}^{n_0}$, the prediction is given by $\hat{y}(\bold x)= \hat{\boldsymbol{\beta}}^\top \, \sigma(\hat{W}\bold x)$. The prediction applied to a set of test data, which was not used for learning $\hat{W}$ and $\hat{\boldsymbol{\beta}}$, gives the test error used for model selection.


\subsection{Algorithm}\label{subsec: algorithm}

In the following, we summarize the proposed method using pseudo-code. We divide the algorithm into two phases. In the first phase (\ref{alg: phase 1}), we compute the matrix $\hat{W}$, which depends on the set of input vectors and the activation function $\sigma(\cdot)$. In the second phase, we compute the output layer weights $\hat{\boldsymbol{\beta}} $ using two possible approaches. In the case of the A-ENR-ELM (\ref{alg: phase 2.a}), we use the matrix $U$ as a proxy for the matrix $\sigma(\hat{W} X)^\top$, and the evaluation of the output layer weights is straightforward. In the case of the I-ENR-ELM (\ref{alg: phase 2.b}), we use the matrix $S = \sigma(\hat{W} X)^\top$ and incrementally add small contributions from each neuron until the residual vector stops improving. Finally, for completeness, we also provide the algorithm for calculating the error curve (\ref{alg: error-evaluation}). When the error curve is evaluated on the test dataset, we obtain the test error curve that can be used to make model selection. Specifically, we propose to choose how many and which neurons to use in the hidden layer of the ENR-ELM by looking at the test error curve.
\begin{algorithm}[ht!]
\renewcommand{\thealgorithm}{algorithm 1}
	\caption{Evaluation of matrix $\hat{W}$}
 \label{alg: phase 1}
	\begin{algorithmic}
      \State
  \State Require $ \sigma:\mathbb R \to\mathbb [-1,1]$ invertible 
		\State Input:  $ X \in \mathbb{R}^{n_0 \times T}$\Comment{$X$ standardized} 
        \State
        \State  \texttt{Evaluate NNGP kernel matrix} $K_{\sigma} \in \mathbb{R}^{T \times T}$ 
		\State  \texttt{Evaluate spectral decomposition} $ K_{\sigma}=U \, \Delta \, U^\top $
		  \State \texttt{Evaluate matrix} $\hat{W}=\sigma^{-1}(U^\top) \, X^\top \, (X X^\top)^{-1}   \in \mathbb{R}^{T \times n_0}$

    \State
    \State \textbf{Return} $\hat{W}$
\end{algorithmic}
\end{algorithm}

 \begin{algorithm}[ht!]
 \renewcommand{\thealgorithm}{algorithm 2}
	\caption{Evaluation of vector $\hat{\boldsymbol{\beta}}$ for  A-ENR-ELM:}
 \label{alg: phase 2.a}
	\begin{algorithmic}
 \State Input $\bold y$, $U$, $n$, $\hat{W}$   \Comment{$y$ centered} 
    \State   \Comment{$n$ maximum number of neurons} 
    \State  \texttt{Initialize}  $\hat{J}=(\,)$
      \For{$l\,=\,1\,:\,n$ }
        \State \texttt{select} $j^*= \text{argmax}_{j \in \hat{J}^c} |< \bold y, U_{.j} >|$
        \State \texttt{update} $\hat{J}=(\hat{J}\,,j^*)$
        \EndFor
        
    \State Evaluate $\hat{\boldsymbol{\beta}}= (U_{\cdot \hat{J}})^\top \bold y$

    \State Update $\hat{W} = \hat{W}_{ \cdot \hat{J}}$
    \State
    \State \textbf{Return} $\hat{W}\,, \hat{\boldsymbol{\beta}} $
    
\end{algorithmic}
\end{algorithm}  
  
 \begin{algorithm}[ht!]
\renewcommand{\thealgorithm}{algorithm 3}
	\caption{ Evaluation of vector $\hat{\boldsymbol{\beta}}$ for  I-ENR-ELM:}
  \label{alg: phase 2.b}
	\begin{algorithmic}
 \State Input  $y$,  $S=\sigma(\hat{W} X)^\top $, $n$, $toll$, $\epsilon$, $\hat{W}$ \Comment{$y$ and $S$ centered} 
    \State     
    \State  \texttt{Initialize} $\hat{\boldsymbol{\beta}}=0 $, $\hat{\bold y}= \bar{\bold y}$, $\bold r=\bold y-\hat{\bold y}  $, $\bold r_{old}=10^{10} \bold{1}$ and $\hat{J}=(\,)$ 
		  \While {$\#(\hat{J}) \leq n$}
      \State \texttt{select} $j^*= \text{argmax}_{j} |< \bold r, S_{.j} >|$
       \State \texttt{update}  $\hat{\beta}_{j^*}= \hat{\beta}_{j^*} + \epsilon < r, S_{.j^*} > / \| S_{.j^*} \|^2  $
        \State \texttt{update}  $\hat{\bold y}=\hat{\bold y} + \epsilon < r, S_{.j^*} > / \| S_{.j^*} \|^2 * S_{.j^*}$
        \State \texttt{update}  $\bold r=\bold r-\epsilon < r, S_{.j^*} > / \| S_{.j^*} \|^2 * S_{.j^*}$
        \State \texttt{if} $j^* \texttt{ not in } \hat{J}$ \texttt{ then } $\hat{J}=(\hat{J}\,,j^*)$
        \State \texttt{if} $(\| \bold r_{old}\| -\bold \| r\| )/ \| \bold y \| < toll $ \texttt{ then break}
        \State \texttt{update} $\bold r_{old} = \bold r$
        \EndWhile
		
    \State Update $\hat{W} = \hat{W}_{\cdot\hat{J}}$
    \State Update $\hat{\boldsymbol{\beta}} = \hat{\boldsymbol{\beta}}_{\hat{J}}$
       \State
    \State \textbf{Return} $\hat{W}\,, \hat{\boldsymbol{\beta}} $
  \end{algorithmic}
\end{algorithm}

\begin{algorithm}[ht!]
 \renewcommand{\thealgorithm}{algorithm 4}
	\caption{ Error curve evaluation}
 \label{alg: error-evaluation}
	\begin{algorithmic}
 \State Input $\bold y$, $X$,$\hat{W} $, $\hat{\boldsymbol{\beta}}$,
 \Comment{$y$ centered by the mean of the training} 
 \State
    \State Evaluate $S=\sigma(\hat{W} X )^\top $\Comment{$X$ standardized by the mean of and std of the training} 
     \State  \texttt{Initialize} $\hat{\bold y}= \bar{\bold y}$
		  \For{$l\,=\,1\,:\,ncol(S)$ }
        \State \texttt{update}  $\hat{\bold y}=\hat{\bold y} + \hat{\beta}_{l} \, S_{.l}$
        \State \texttt{evaluate error} $Err(l)= \|\bold y - \hat{\bold y}\|  / \| \bold y\|$
        \EndFor 
\end{algorithmic}
\end{algorithm}  

Regarding the second phase of the proposed method, the choice of the parameter $n$, the maximum number of columns (i.e. the maximum number of neurons) depends on the dataset size. Notably, $n >> n_0$ since the dimension of the feature map's target space is expected to exceed that of its input space. Similarly, in data-rich scenarios, where $T$ is significantly large, constructing error curves as the number of neurons approaches $T$ becomes unnecessary. 
%
It needs to be noted that both approaches assign importance to the columns of $\hat{W}^\top$ and rearrange them accordingly. Then, the first approach (A-ENR-ELM) considers neurons up to the maximum number $n$; while the second approach (I-ENR-ELM) saturates the number of neurons when the convergence criterion is met. In the respective algorithms, the ordered set $\hat{J}$ enables this rearrangement and selection mechanism.

Concerning the I-ENR-ELM, the choice of the parameter $\epsilon$ is critical, with  $\epsilon \in (0,1]$. When $\epsilon=1$, the algorithm progresses with larger steps, following the classical forward stagewise regression. Conversely, when  $\epsilon <1$, the steps are smaller, resulting in slower convergence. However, as noted in \citep{hastie2009elements} cf Section 3.3.3 ``this \emph{slow fitting} can pay dividends in high-dimensional problems". 

We stress that the output parameter $\hat{W}$ and $\hat{\boldsymbol{\beta}}$ returned by the algorithm do not represent those of a single ENR-ELM, but rather represent the parameters of a nested family of ENR-ELMs that differ by one hidden layer neuron each. This is extremely important since it allows the construction of the test error curve, as explained in the \ref{alg: error-evaluation}, speeding up the entire model selection procedure. 
On the other hand, for traditional ELMs, the test error curve evaluation is performed by evaluating for each fixed number of neurons one or more realizations of the ELM, requiring a computational cost several orders of magnitude higher than the one we propose. This difference is highlighted by the experimental results presented in the next section.

\section{Numerical study}\label{sec: experimental}

In this section, we present numerical experiments performed on synthetic and real datasets to compare traditional ELM and the proposed ENR-ELM methods in terms of prediction accuracy and computational time.
We ran all experiments on a computer with the specification given in Table \ref{table-computer-specification}. All the results are reproducible using the code available at \emph{https://github.com/FabianoVeglianti/Effective-Non-Random-Extreme-Learning-Machine}. All experiments were implemented in Python 3.10; the required libraries are specified in the code repository.

\begin{table}[hbp]
\caption{Specification of the computer on which the experiments were run}\label{table-computer-specification}%
\begin{tabular}{@{}llll@{}}
\toprule
Parameter & Value\\
\midrule
Processor & Intel i5-9300H \\
RAM size & 8 GB \\
GPU & NVidia GeForce GTX 1650 \\
Operating system & Windows 11 Home \\
\botrule
\end{tabular}
\end{table}

\subsection{Experimental setting}
For each dataset considered in the experiments, we applied a random splitting into a training set (75\% of the data) and a test set (25\% of the data). To ensure comparability between features, we applied standardization to the training set by removing the mean and scaling the data to unit variance. The mean and the standard deviation of the training dataset were used to standardize the test set before evaluating the prediction accuracy.

\noindent

We applied the algorithms proposed in Section \ref{subsec: algorithm} to the training data to obtain both the A-ENR-ELM model parameters and the I-ENR-ELM model parameters; after that, we used these parameters to compute training and test error curves by the procedure described in \ref{alg: error-evaluation}. 
The total computational time for the proposed ENR-ELM models includes the NNGP kernel matrix evaluation, its spectral decomposition, and the training and test error curves evaluation.

\noindent

For the sake of comparison, we performed traditional ELM on the same datasets, and for each fixed number of hidden layer neurons, we evaluated 20 independent realizations of the ELM. For each run, we recorded the training and test errors and reported the mean of these values across the 20 runs. In terms of computational time, we measured the total time required for all 20 runs as well as the average time per individual run.

\noindent

To account for fluctuations in the computational time measurements for model selection, which may be caused by background processes independent of human activities, the procedure was repeated 10 times. The mean elapsed time across these repetitions is reported to provide a more robust estimate.

The model selection process involves the evaluation of ELM, A-ENR-ELM, and I-ENR-ELM, with the number of hidden layer units set to a maximum value of $n = \min{\{50 \times n_0, \#Training/2\}}$. This choice is motivated by two considerations: first, it ensures that the hidden layer can project the data into a significantly larger space than the input space, facilitating a more expressive feature transformation; second, it guarantees that the resulting minimization given in eq.~\eqref{eq: betahat_defmatrix} remains well-conditioned, even for datasets with limited sample sizes, thus contributing to the numerical stability of the training process.

The training and test errors were measured with the normalized Root Mean Squared Error (RMSE) metric given in the following expression
\begin{equation}\label{eq: RMSE}
     \| \bold y-  \hat{\bold y} \| / \| \bold y \| = \sqrt{ \sum_{i=1}^{T} (y_i - \hat{y}_i)^2 /   \sum_{i=1}^{T}  y_i^2}  
\end{equation}
where $\mathbf{y}$ represents the true target values, $\mathbf{\hat{y}}$ denotes the predicted target values, and T is the number of samples.
 We normalize the RMSE by the L2-norm of the true target variable to enable comparisons across datasets with different target variable magnitudes. This ensures that the error magnitude remains independent of the target variable scale.

\subsection{Results on synthetic datasets}

Each synthetic dataset was created by generating $T \in \{300, 1200\}$ pairs $\{(\bold x_l,y_l)\}_{l=1,...,T}$ where $\bold x_l \in  \mathbb R^{n_0}$ with dimension $n_0=20$ or $n_0=80$ and $y_l=f(\bold x_l)+ \epsilon_l$, with $f(\cdot)$ being a given function and $\epsilon_l$ are independent and identically distributed Gaussian variables with zero mean and  fixed variance $\sigma^2$. The generic input data vectors $\bold x$ were independently sampled from the following three different types of distribution: \emph{Uniform}, where the vector components are independent and uniformly distributed in the interval $[- 2 \pi, 2 \pi]$;  \emph{Indep. Gaussian}, where the vector components are independent and identically distributed as a standard Gaussian variable; \emph{Toepl. Gaussian},  where the vector $\bold x$ is distributed as $n_0$-variate Gaussian variable with zero mean and covariance given by  $cov(x_i,x_j)=\rho^{|i-j|}$ with $\rho = 0.8$. We consider two types of function $f(\cdot)$: \emph{Linear}, where. $f( \bold x)=\bold \alpha^\top \bold x + \beta$ where the components of $\bold \alpha$ and $\beta$ are independent and uniformly distributed in the interval $[- 2, 2]$, and \emph{Shallow NN}, where $f( \bold x)$ is the output obtained by a Neural Network with one hidden layer composed of 100 neurons whose hidden layer weights are drawn from a Gaussian distribution with zero mean and variance $1/n_0$, while the output layer weights are generated according to a Gaussian distribution with zero mean and variance $1/100$. Finally, we fix the variance of the noise $\sigma^2$ such that the Signal to Noise Ratio (SNR), defined as the ratio between the variance of the unknown signal and the variance of the noise, results to be $SNR=2$ or $SNR=10$, the first case representing a strong noise case, the second case representing a weak noise case. 

\noindent

With the two different choices of the sample size ($T=300$ or $T=1200$), two different choices of the input dimension ($n_0=20$ or $n_0=80$), the three different choices of input data generation (\emph{Uniform} or \emph{Indep. Gaussian} or \emph{Toepl. Gaussian}), the two different choices of unknown function (\emph{Linear}, \emph{Shallow NN}) and the two different choices of SNR ($SNR=2$ or $SNR=10$) we have a total of 48 different datasets, which are summarized in Table \ref{table-datasets-sintetici}. 

\begin{figure}[!h]
\centering
\includegraphics[width=\textwidth,height=\textheight,keepaspectratio]{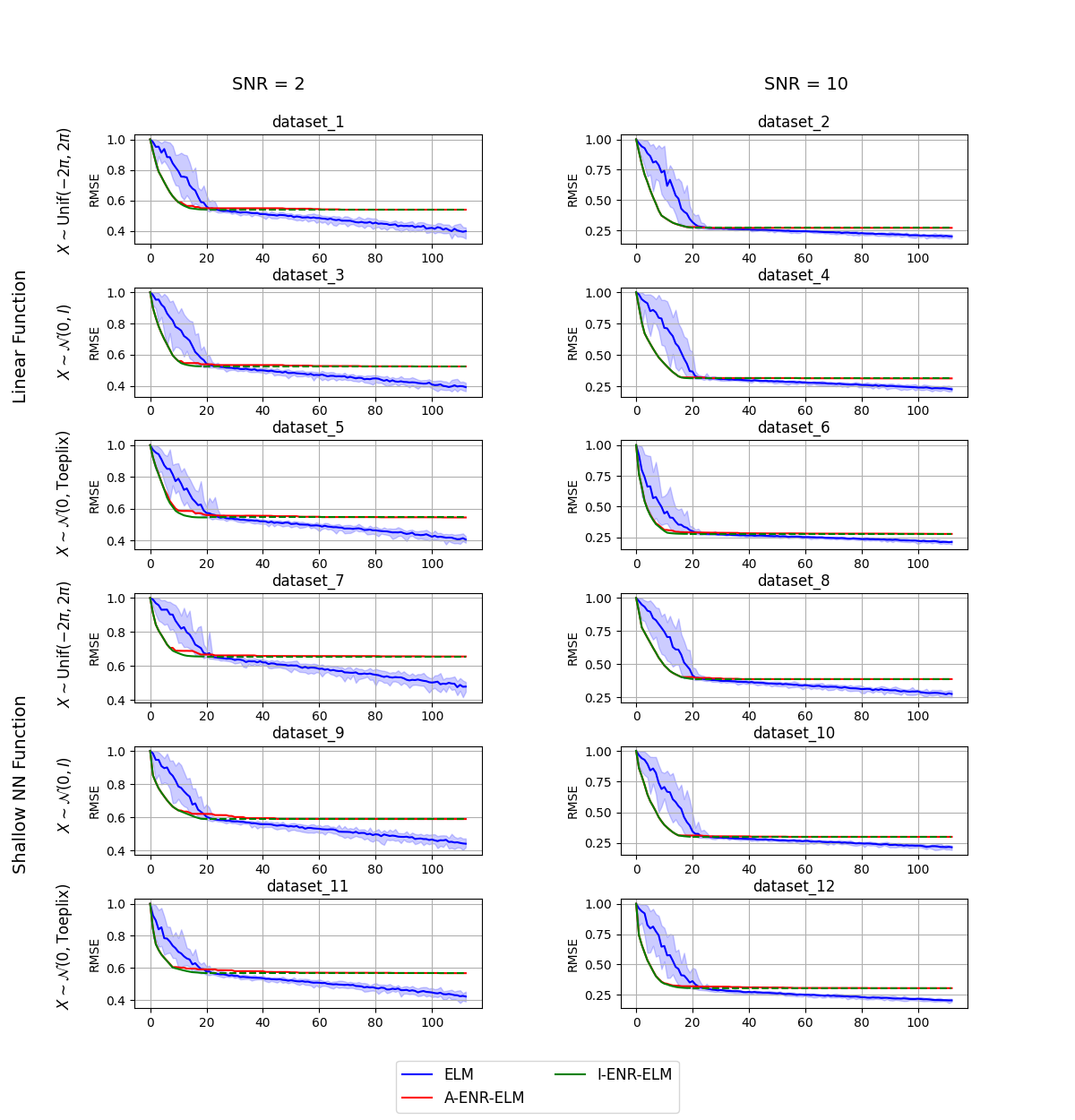}
\caption{Training error of ELM, incremental-ENR-ELM, and approximated-ENR-ELM on synthetic datasets with input dimension $n_0=20$. The ELM curve represents the mean training error over 20 independent realizations of the ELM model, with the shaded region indicating the range between the minimum and maximum values observed across these realizations.}\label{fig: synthetic_datasets_training_300_20}
\end{figure}

\begin{table}[!h]
\caption{Characteristics of synthetic datasets}\label{table-datasets-sintetici}
\begin{tabular}{@{}lllllll@{}}
\toprule
Dataset & $T$ & $n_0$ & Input distribution & $f(\cdot)$ & SNR \\ 
\midrule
Dataset 1  & 300  & 20 & Uniform         & Linear     & 2  \\
Dataset 2  & 300  & 20 & Uniform         & Linear     & 10 \\
Dataset 3  & 300  & 20 & Indep. Gaussian & Linear     & 2  \\
Dataset 4  & 300  & 20 & Indep. Gaussian & Linear     & 10 \\
Dataset 5  & 300  & 20 & Toepl. Gaussian & Linear     & 2  \\
Dataset 6  & 300  & 20 & Toepl. Gaussian & Linear     & 10 \\
Dataset 7  & 300  & 20 & Uniform         & Shallow NN & 2  \\
Dataset 8  & 300  & 20 & Uniform         & Shallow NN & 10 \\
Dataset 9  & 300  & 20 & Indep. Gaussian & Shallow NN & 2  \\
Dataset 10 & 300  & 20 & Indep. Gaussian & Shallow NN & 10 \\
Dataset 11 & 300  & 20 & Toepl. Gaussian & Shallow NN & 2  \\
Dataset 12 & 300  & 20 & Toepl. Gaussian & Shallow NN & 10 \\
Dataset 13 & 300  & 80 & Uniform         & Linear     & 2  \\
Dataset 14 & 300  & 80 & Uniform         & Linear     & 10 \\
Dataset 15 & 300  & 80 & Indep. Gaussian & Linear     & 2  \\
Dataset 16 & 300  & 80 & Indep. Gaussian & Linear     & 10 \\
Dataset 17 & 300  & 80 & Toepl. Gaussian & Linear     & 2  \\
Dataset 18 & 300  & 80 & Toepl. Gaussian & Linear     & 10 \\
Dataset 19 & 300  & 80 & Uniform         & Shallow NN & 2  \\
Dataset 20 & 300  & 80 & Uniform         & Shallow NN & 10 \\
Dataset 21 & 300  & 80 & Indep. Gaussian & Shallow NN & 2  \\
Dataset 22 & 300  & 80 & Indep. Gaussian & Shallow NN & 10 \\
Dataset 23 & 300  & 80 & Toepl. Gaussian & Shallow NN & 2  \\
Dataset 24 & 300  & 80 & Toepl. Gaussian & Shallow NN & 10 \\
Dataset 25 & 1200 & 20 & Uniform         & Linear     & 2  \\
Dataset 26 & 1200 & 20 & Uniform         & Linear     & 10 \\
Dataset 27 & 1200 & 20 & Indep. Gaussian & Linear     & 2  \\
Dataset 28 & 1200 & 20 & Indep. Gaussian & Linear     & 10 \\
Dataset 29 & 1200 & 20 & Toepl. Gaussian & Linear     & 2  \\
Dataset 30 & 1200 & 20 & Toepl. Gaussian & Linear     & 10 \\
Dataset 31 & 1200 & 20 & Uniform         & Shallow NN & 2  \\
Dataset 32 & 1200 & 20 & Uniform         & Shallow NN & 10 \\
Dataset 33 & 1200 & 20 & Indep. Gaussian & Shallow NN & 2  \\
Dataset 34 & 1200 & 20 & Indep. Gaussian & Shallow NN & 10 \\
Dataset 35 & 1200 & 20 & Toepl. Gaussian & Shallow NN & 2  \\
Dataset 36 & 1200 & 20 & Toepl. Gaussian & Shallow NN & 10 \\
Dataset 37 & 1200 & 80 & Uniform         & Linear     & 2  \\
Dataset 38 & 1200 & 80 & Uniform         & Linear     & 10 \\
Dataset 39 & 1200 & 80 & Indep. Gaussian & Linear     & 2  \\
Dataset 40 & 1200 & 80 & Indep. Gaussian & Linear     & 10 \\
Dataset 41 & 1200 & 80 & Toepl. Gaussian & Linear     & 2  \\
Dataset 42 & 1200 & 80 & Toepl. Gaussian & Linear     & 10 \\
Dataset 43 & 1200 & 80 & Uniform         & Shallow NN & 2  \\
Dataset 44 & 1200 & 80 & Uniform         & Shallow NN & 10 \\
Dataset 45 & 1200 & 80 & Indep. Gaussian & Shallow NN & 2  \\
Dataset 46 & 1200 & 80 & Indep. Gaussian & Shallow NN & 10 \\
Dataset 47 & 1200 & 80 & Toepl. Gaussian & Shallow NN & 2  \\
Dataset 48 & 1200 & 80 & Toepl. Gaussian & Shallow NN & 10 \\
\botrule
\end{tabular}
\end{table}
Fig. \ref{fig: synthetic_datasets_training_300_20} shows the training error of the ELM, A-ENR-ELM, and I-ENR-ELM models across the first 12 synthetic datasets with sample size $T=300$ and input dimension $n_0 = 20$ in terms of normalized root mean square error (RMSE). These results indicate that the proposed models avoid overfitting effectively. Specifically, for the A-ENR-ELM model, the training error saturates beyond a certain point, reflecting its ability to prevent overfitting. Similarly, the I-ENR-ELM model shows a plateau in training error upon reaching its stopping criterion, thereby preventing the addition of unnecessary neurons. By contrast, the traditional ELM model tends to fit all available data, potentially leading to interpolation and overfitting.

\begin{figure}[!h]
\centering
\includegraphics[width=\textwidth,height=\textheight,keepaspectratio]{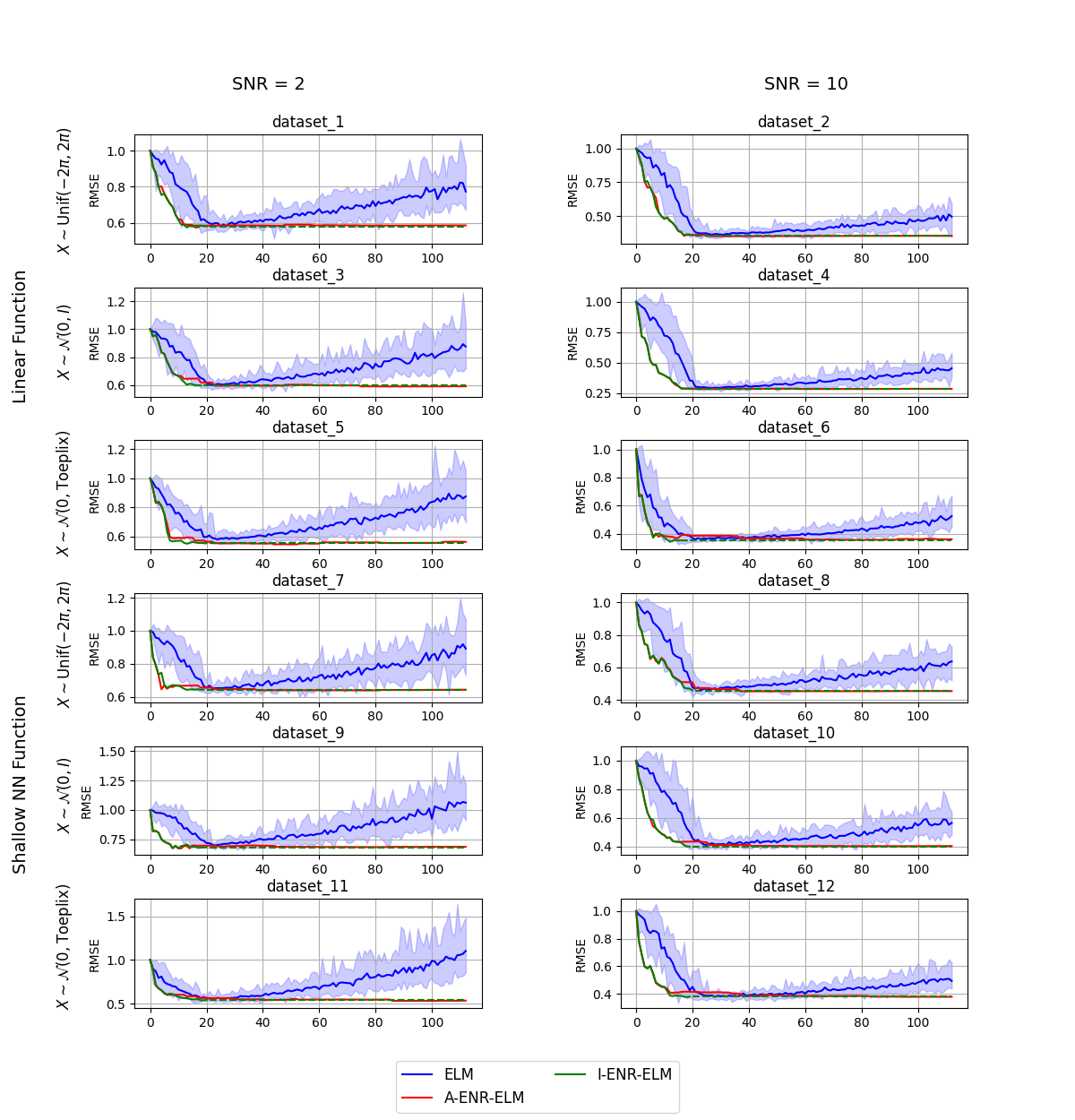}
\caption{Test error of ELM, incremental-ENR-ELM, and approximated-ENR-ELM on synthetic datasets with input dimension $n_0=20$. The ELM curve represents the mean test error over 20 independent realizations of the ELM model, with the shaded region indicating the range between the minimum and maximum values observed across these realizations.}\label{fig: synthetic_datasets_test_300_20}
\end{figure}

Fig. \ref{fig: synthetic_datasets_test_300_20} presents the test error curves for the same models. This figure further corroborates the tendency of the traditional ELM to overfit the data, whereas both the A-ENR-ELM and I-ENR-ELM models demonstrate resilience against overfitting. Moreover, Fig. \ref{fig: synthetic_datasets_test_300_20} highlights that the proposed models converge more rapidly to the minimum test error, and achieve performance comparable to that of the traditional ELM. In Fig. \ref{fig: synthetic_datasets_training_300_20} and Fig. \ref{fig: synthetic_datasets_test_300_20} the I-ENR-ELM curve transitions from solid to dashed at the point where the convergence criterion is met. Beyond this point, the last error value is replicated and shown as a dashed line to facilitate comparison with the other curves.

Table \ref{table: performance-synthetic-1-12} presents the minimum observed test error achieved by each model across the first 12 synthetic datasets and the associated number of hidden layer units required. The results reveal that the I-ENR-ELM and traditional ELM models exhibit similar hidden layer neuron requirements. However, the A-ENR-ELM model generally demands more hidden units to achieve its minimum test error. Notably, even though the A-ENR-ELM model is highly resistant to overfitting, it continues to progressively fit the information in the data as the number of neurons increases. This continued adaptation leads to a steady decrease in training error, although marginally, as more neurons are added. This behavior causes the test error to fluctuate around a minimum value, typically reached early in the training process. As a result, using the minimum test error to determine the optimal number of neurons for the A-ENR-ELM model is not a reliable criterion, as this approach often yields an inflated estimate of the required number of neurons due to these test error fluctuations. A more appropriate approach for this model could be choosing the point at which the test error curve has the greatest curvature. Selecting the point where the test error curve shows the greatest curvature is generally the most suitable approach for all methods. Indeed, examining the shapes of the test error curves, we see that the minimum point reported in Table \ref{table: performance-synthetic-1-12}  is less parsimonious than the point of maximum curvature identifiable in Fig. \ref{fig: synthetic_datasets_test_300_20}.
\noindent

So far, we have used only 12 of the 48 synthetic datasets. Similar conclusions can be drawn by analyzing the results obtained in the remaining synthetic datasets, see appendix \ref{secA1}.
 
Table \ref{table: computational-time-synthetic}
contains the computational time required for all considered methods and 48 synthetic datasets. The results shown in Table \ref{table: computational-time-synthetic} demonstrate that the proposed methods exhibit greater computational efficiency compared to the traditional ELM.  Notably, the A-ENR-ELM proves to be significantly more computationally efficient. This comparison also highlights the key factors influencing the computational time of ENR-ELM algorithms: sample size and input dimensionality. Specifically, sample size predominantly determines the computational time for A-ENR-ELM, as observed from the almost constant computational time for datasets with sample size $T=300$, followed by a marked increase for larger datasets with sample size $T=1200$. This indicates that, for A-ENR-ELM, the primary computational cost arises from the eigen decomposition of the matrix $K_{\sigma}$. While this cost is also incurred by the I-ENR-ELM model, for this model also input dimensionality plays a prominent role. Finally, input dimensionality seems to be almost negligible for the time required by both A-ENR-ELM and ELM models.
The results further indicate that the computational time for an exhaustive model selection using the traditional ELM algorithm increases with larger sample sizes. This increase can be attributed to the exhaustive nature of the model selection process, where the performance is evaluated across a range of hidden layer units from 1 to $n = \min{\{50 \times n_0, \#Training/2\}}$. Although a reduced number of hidden units could be considered for ELM to expedite the model selection process, doing so would introduce bias in the comparison, favoring ELM in terms of computational time while simultaneously disadvantaging it in terms of predictive accuracy. Therefore, maintaining consistency in model selection ensures a fair and comprehensive comparison.

\begin{table}[!h]
\caption{Optimal number of hidden layer neurons and relative test error (the minimum of the test error curve) for synthetic datasets with sample size $T=300$ and input dimension $n_0 = 20$. For the traditional ELM the minimum is evaluated on the mean test error curve across 20 realizations and its standard deviation is reported.}
\label{table: performance-synthetic-1-12}
\begin{tabular}{l|cc|cc|cc}
\toprule
\multirow{2}{*}{\textbf{dataset}} & \multicolumn{2}{c|}{\textbf{A-ENR-ELM}} & \multicolumn{2}{c|}{\textbf{I-ENR-ELM}} & \multicolumn{2}{c}{\textbf{ELM}} \\
 & \textbf{n} & \textbf{value} & \textbf{n} & \textbf{value} & \textbf{n} & \textbf{value} \\
\midrule
Dataset 1  & 15  & 0.5738  & 13  & 0.5741  & 24  & 0.5805 $\pm$ 0.0171 \\
Dataset 2  & 32  & 0.3509  & 20  & 0.3562  & 30  & 0.3612 $\pm$ 0.0104 \\
Dataset 3  & 85  & 0.5898  & 18  & 0.5976  & 21  & 0.5996 $\pm$ 0.0170 \\
Dataset 4  & 61  & 0.2849  & 20  & 0.2871  & 26  & 0.2939 $\pm$ 0.0134 \\
Dataset 5  & 51  & 0.5463  & 13  & 0.5511  & 23  & 0.5799 $\pm$ 0.0179 \\
Dataset 6  & 76  & 0.3568  & 12  & 0.3439  & 22  & 0.3616 $\pm$ 0.0083 \\
Dataset 7  & 75  & 0.6379  & 13  & 0.6418  & 21  & 0.6509 $\pm$ 0.0107 \\
Dataset 8  & 39  & 0.4526  & 20  & 0.4579  & 22  & 0.4606 $\pm$ 0.0172 \\
Dataset 9  & 10  & 0.6732  & 10  & 0.6732  & 22  & 0.6973 $\pm$ 0.0134 \\
Dataset 10 & 79  & 0.4021  & 19  & 0.3991  & 31  & 0.4104 $\pm$ 0.0184 \\
Dataset 11 & 92  & 0.5298  & 18  & 0.5362  & 19  & 0.5565 $\pm$ 0.0192 \\
Dataset 12 & 96  & 0.3775  & 18  & 0.3761  & 27  & 0.3770 $\pm$ 0.0097 \\
\bottomrule
\end{tabular}
\end{table}

\begin{table}[!h]
\caption{Computational time on synthetic datasets expressed in milliseconds.}\label{table: computational-time-synthetic}
\centering
\begin{tabular}{@{}lcccc@{}}
\toprule
\textbf{Dataset} & \textbf{A-ENR-ELM} & \textbf{I-ENR-ELM} & \textbf{ELM (20 iterations)} & \textbf{ELM (1 iteration)} \\ 
\midrule
Dataset 1  & 16.32146   & 124.96869  & 5307.37014   & 265.36851   \\
Dataset 2  & 14.18106   & 140.60400  & 3787.67999   & 189.38400   \\
Dataset 3  & 14.54270   & 117.02895  & 3738.63963   & 186.93198   \\
Dataset 4  & 14.25233   & 121.38743  & 4370.76972   & 218.53849   \\
Dataset 5  & 14.08346   & 105.09719  & 3819.09180   & 190.95459   \\
Dataset 6  & 14.75797   & 107.49984  & 3645.48390   & 182.27419   \\
Dataset 7  & 15.93076   & 115.34181  & 3760.09064   & 188.00453   \\
Dataset 8  & 13.86249   & 114.22304  & 3777.84259   & 188.89213   \\
Dataset 9  & 14.53062   & 94.65098   & 3748.16894   & 187.40845   \\
Dataset 10 & 13.96850   & 143.14260  & 3757.50901   & 187.87545   \\
Dataset 11 & 14.24737   & 106.52110  & 6491.89338   & 324.59467   \\
Dataset 12 & 14.36173   & 136.39106  & 5169.10815   & 258.45541   \\
Dataset 13 & 20.54475   & 333.03524  & 6069.08772   & 303.45439   \\
Dataset 14 & 24.03600   & 435.90489  & 5367.00658   & 268.35033   \\
Dataset 15 & 16.05688   & 335.69949  & 4970.70184   & 248.53509   \\
Dataset 16 & 16.03769   & 408.20349  & 4678.19782   & 233.90989   \\
Dataset 17 & 15.10580   & 278.86309  & 4474.52763   & 223.72638   \\
Dataset 18 & 18.08123   & 403.01774  & 4430.76436   & 221.53822   \\
Dataset 19 & 15.46903   & 312.33403  & 4566.51599   & 228.32580   \\
Dataset 20 & 15.08143   & 369.08719  & 4375.69521   & 218.78476   \\
Dataset 21 & 17.08805   & 329.02546  & 4371.82896   & 218.59145   \\
Dataset 22 & 16.03508   & 370.74961  & 4475.03478   & 223.75174   \\
Dataset 23 & 15.17585   & 262.05518  & 4395.86727   & 219.79336   \\
Dataset 24 & 15.59896   & 300.90870  & 4304.27108   & 215.21355   \\
Dataset 25 & 338.16093  & 687.32146  & 135352.25036 & 6767.61252  \\
Dataset 26 & 325.93514  & 748.87876  & 130395.48539 & 6519.77427  \\
Dataset 27 & 335.48370  & 764.41084  & 130706.77772 & 6535.33889  \\
Dataset 28 & 329.64922  & 715.07219  & 131380.34171 & 6569.01709  \\
Dataset 29 & 321.74117  & 595.92242  & 132626.76983 & 6631.33849  \\
Dataset 30 & 326.42065  & 730.79328  & 128876.64502 & 6443.83225  \\
Dataset 31 & 331.72239  & 666.00263  & 129583.91208 & 6479.19560  \\
Dataset 32 & 321.78727  & 692.16751  & 130419.67952 & 6520.98398  \\
Dataset 33 & 329.06390  & 717.21019  & 130369.22295 & 6518.46115  \\
Dataset 34 & 328.76441  & 697.19879  & 131257.85343 & 6562.89267  \\
Dataset 35 & 330.05414  & 579.63691  & 132284.08713 & 6614.20436  \\
Dataset 36 & 325.93300  & 694.37587  & 150845.70616 & 7542.28531  \\
Dataset 37 & 357.21692  & 1448.53690 & 140409.61809 & 7020.48090  \\
Dataset 38 & 345.47655  & 1598.43018 & 138844.50821 & 6942.22541  \\
Dataset 39 & 344.23160  & 1365.46915 & 140113.51526 & 7005.67576  \\
Dataset 40 & 349.53607  & 1652.66787 & 142978.33630 & 7148.91681  \\
Dataset 41 & 342.46273  & 1050.00356 & 138376.15035 & 6918.80752  \\
Dataset 42 & 344.69478  & 1194.74726 & 139728.52306 & 6986.42615  \\
Dataset 43 & 347.03367  & 1358.38221 & 137371.92784 & 6868.59639  \\
Dataset 44 & 350.87913  & 1350.14249 & 141273.82691 & 7063.69135  \\
Dataset 45 & 346.11636  & 1254.28499 & 141724.36133 & 7086.21807  \\
Dataset 46 & 358.33137  & 1608.32152 & 138566.31555 & 6928.31578  \\
Dataset 47 & 342.84621  & 1152.46139 & 140474.07547 & 7023.70377  \\
Dataset 48 & 353.20407  & 1135.06300 & 139754.34675 & 6987.71734  \\
\botrule
\end{tabular}
\end{table}

\subsection{Results on real datasets}

The datasets used in this section were acquired from the UCI \cite{UCIRepository} repository, from the dataset repository of The Elements of Statistical Learning\cite{hastie2009elements}\footnote{\href{https://hastie.su.domains/ElemStatLearn/data.html}{https://hastie.su.domains/ElemStatLearn/data.html}} and from the University of Porto repository\footnote{\href{https://www.dcc.fc.up.pt/~ltorgo/Regression/}{https://www.dcc.fc.up.pt/~ltorgo/Regression/}}. Table \ref{table-datasets-reali} contains the characteristics of the datasets along with their description.

\begin{table}[!h]
\caption{Characteristics and description of real datasets. The column "number of features" ($n_0$) represents the total feature count after applying one-hot encoding to categorical variables. The code used for this encoding is available in the code repository.}\label{table-datasets-reali}
\renewcommand{\arraystretch}{1.4}
\begin{tabular}{@{}p{12.5mm}p{12.5mm}p{12.5mm}p{13.5mm}p{14.5mm}p{54.5mm}@{}}
\toprule
Dataset & $T$ & $n_0$ & Min. value of $y$ & Max. value of $y$ & Description \\ 
\midrule
Abalone & 4177 & 10 & 1.0 & 29.0 & Predicting the age of abalone from physical measurements. \\
Auto MPG & 392 & 21 & 9.0 & 46.6 & The data concerns city-cycle fuel consumption in miles per gallon, to be predicted in terms of 3 multivalued discrete and 5 continuous attributes. \\
California Housing & 20640 & 8 & 0.15 & 5.0 & This dataset was derived from the 1990 U.S. census, using one row per census
block group. The goal is to predict the median house value. \\
Delta Ailerons & 7129 & 5 & -0.0021 & 0.0022 & Prediction of variance of ailerons positioning in the F-16 fighter aircraft. \\
LA Ozone & 330 & 9 & 1.0 & 38.0 & These data record the level of atmospheric ozone concentration from
eight daily meteorological measurements made in the Los Angeles basin
in 1976. The goal is to predict the upland maximum ozone. \\
Machine CPU & 209 & 6 & 6.0 & 1550.0 & Relative CPU Performance Data, described in terms of its cycle time, memory size, etc. The goal is to predict the CPU performance. \\
Prostate Cancer & 97 & 8 & -0.4308 & 5.5829 & Clinical data for men who have undergone a prostatectomy. The goal is to predict the logarithm of the prostate-specific antigen (PSA) level. \\
Servo & 167 & 19 & 0.1312 & 7.1 & Predicting the rise time of a servomechanism in terms of two (continuous) gain settings and two (discrete) choices of mechanical linkages. \\
\botrule
\end{tabular}
\end{table}

\begin{figure}[!h]
\centering
\includegraphics[width=\textwidth,height=\textheight,keepaspectratio]{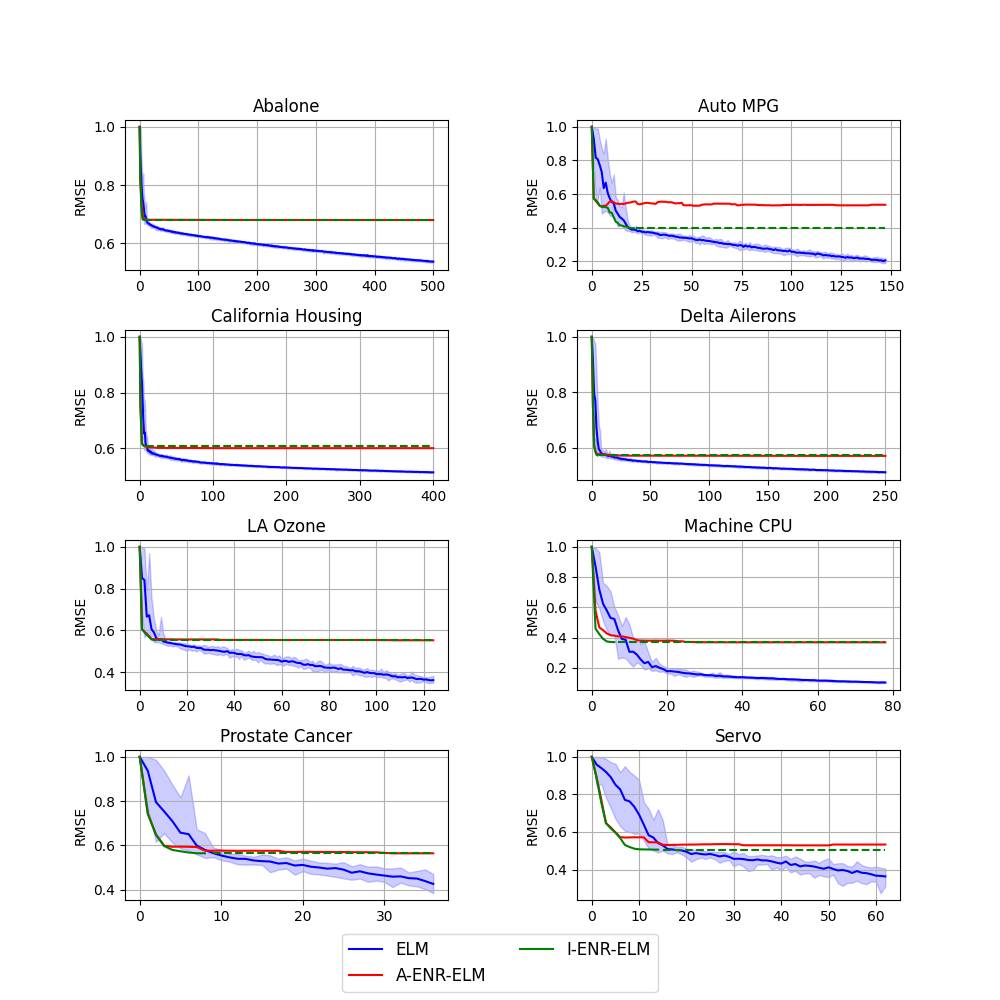}
\caption{Training error of ELM, incremental-ENR-ELM, and approximated-ENR-ELM on real datasets. The ELM curve represents the mean training error over 20 independent realizations of the ELM model, with the shaded region indicating the range between the minimum and maximum values observed across these realizations.}\label{fig: real-datasets-training}
\end{figure}

Fig. \ref{fig: real-datasets-training} illustrates the training error of the ELM, A-ENR-ELM, and I-ENR-ELM models across real datasets. The results on real datasets align with those observed on synthetic datasets: both A-ENR-ELM and I-ENR-ELM successfully avoid overfitting, whereas the traditional ELM exhibits overfitting. Additionally, the results reveal that for the Auto MPG dataset, the A-ENR-ELM model exhibits a higher saturation point in training error compared to I-ENR-ELM, suggesting that these two models may perform differently depending on the dataset.

\begin{figure}[!h]
\centering
\includegraphics[width=\textwidth,height=\textheight,keepaspectratio]{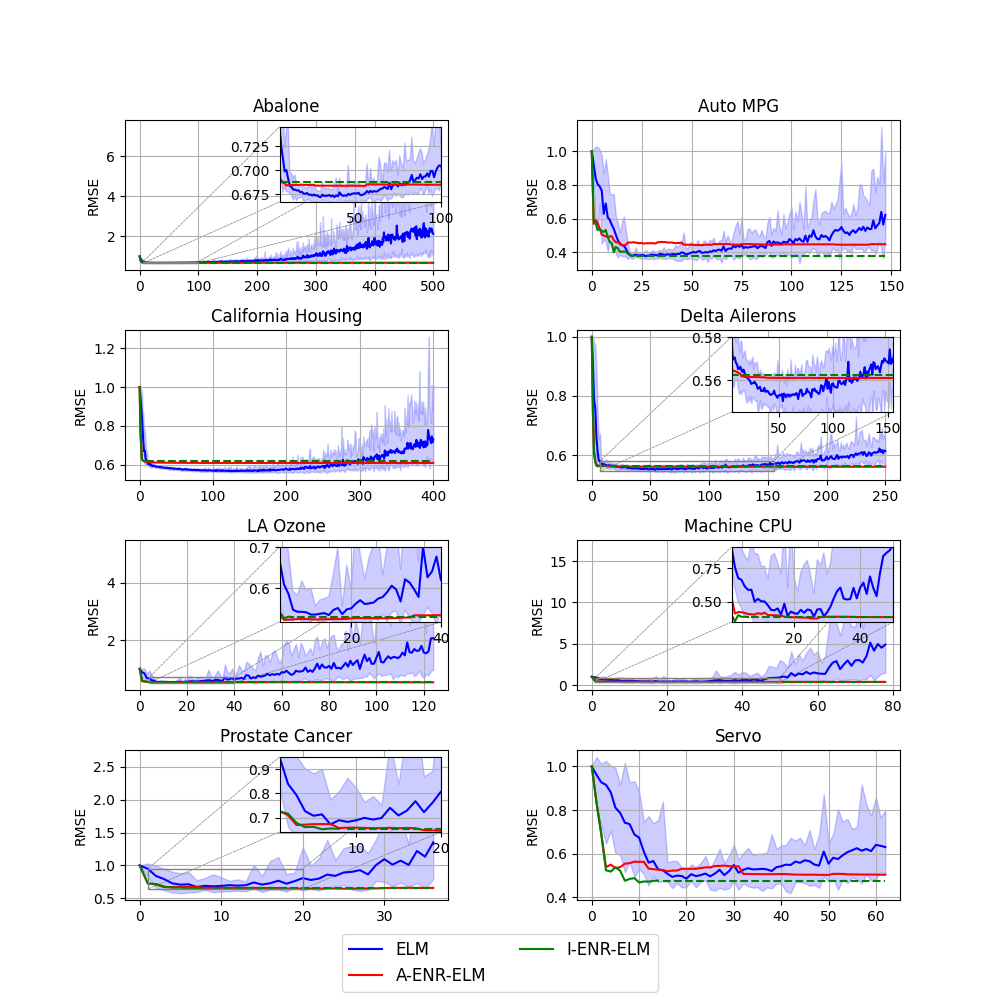}
\caption{Test error of ELM, incremental-ENR-ELM, and approximated-ENR-ELM on real datasets. The ELM curve represents the mean test error over 20 independent realizations of the ELM model, with the shaded region indicating the range between the minimum and maximum values observed across these realizations.}\label{fig: real-datasets-test}
\end{figure}

Fig. \ref{fig: real-datasets-test} depicts the test error curves of the ELM, A-ENR-ELM, and I-ENR-ELM models across real datasets. This figure further confirms that the proposed methods converge more rapidly to the minimum error than the traditional ELM. In terms of overall predictive performance, the ENR-ELM methods demonstrate comparable generalization capabilities to the traditional ELM, with performance varying slightly across datasets. However, consistently with the observation on the training error curve, in the case of the Auto MPG dataset, the A-ENR-ELM model performs differently from the I-ENR-ELM model.

Table \ref{table: performance-real} shows the minimum test error achieved by each model across all real datasets, along with the number of hidden layer units required to reach this minimum. Unlike the results obtained on synthetic datasets, the I-ENR-ELM model attains the minimum error using fewer hidden units than the traditional ELM, highlighting its greater parsimony. On the other hand, as observed on synthetic datasets, the A-ENR-ELM model requires a greater number of hidden layer units to achieve the minimum error. Again, this is because the best criterion for the A-ENR-ELM model should be selecting the point of greatest curvature in the test error curve rather than its minimum point. 

 Table \ref{table:computational-time-real} contains the computational time required for model selection. The results support the findings from experiments on synthetic datasets. Notably, it reveals that the proposed models offer no substantial computational efficiency benefit over a single instance of the traditional ELM when the training set is large. In particular, when  $T > 5000$, as with the California Housing and Delta Ailerons datasets, there is even a decrease in efficiency. This is largely due to the high computational cost associated with the eigen-decomposition of the matrix $K_{\sigma} \in \mathbb{R}^{T \times T}$. In principle, this cost could potentially be mitigated by employing subsampling of the training dataset.


\begin{table}[!h]
\caption{Optimal number of hidden layer neurons and relative test error (the minimum of the test error curve). For the traditional ELM the minimum is evaluated on the mean test error curve across 20 realizations and its std is reported.}
\label{table: performance-real}
\centering
\begin{tabular}{l|cc|cc|cc}
\toprule
\multirow{2}{*}{\textbf{dataset}} & \multicolumn{2}{c|}{\textbf{A-ENR-ELM}} & \multicolumn{2}{c|}{\textbf{I-ENR-ELM}} & \multicolumn{2}{c}{\textbf{ELM}} \\
 & \textbf{n} & \textbf{value} & \textbf{n} & \textbf{value} & \textbf{n} & \textbf{value} \\
\midrule
Abalone            & 51  & 0.6839  & 6   & 0.6878  & 29  & 0.6721 $\pm$ 0.0035 \\
Auto MPG           & 14  & 0.4422  & 21  & 0.3789  & 26  & 0.3771 $\pm$ 0.0120 \\
California Housing & 387 & 0.6080  & 7   & 0.6160  & 160 & 0.5653 $\pm$ 0.0050 \\
Delta Ailerons     & 248 & 0.5608  & 5   & 0.5622  & 53  & 0.5501 $\pm$ 0.0052 \\
LA Ozone           & 4   & 0.5231  & 5   & 0.5269  & 14  & 0.5350 $\pm$ 0.0129 \\
Machine CPU        & 25  & 0.3751  & 2   & 0.3543  & 25  & 0.3899 $\pm$ 0.0764 \\
Prostate Cancer    & 24  & 0.6465  & 6   & 0.6530  & 6   & 0.6729 $\pm$ 0.0456 \\
Servo              & 47  & 0.5031  & 10  & 0.4685  & 19  & 0.4860 $\pm$ 0.0203 \\
\bottomrule
\end{tabular}
\end{table}

\begin{table}[!h]
\caption{Computational time on real datasets expressed in milliseconds.}\label{table:computational-time-real}
\centering
\begin{tabular}{@{}lcccc@{}}
\toprule
\textbf{Dataset} & \textbf{A-ENR-ELM} & \textbf{I-ENR-ELM} & \textbf{ELM (20 iterations)} & \textbf{ELM (1 iteration)} \\ 
\midrule
Abalone            & $1.03 \times 10^4$    & $1.57 \times 10^4$     & $6.45 \times 10^5$            & $3.22 \times 10^4$      \\
Auto MPG           & $2.70 \times 10^1$    & $1.98 \times 10^2$     & $9.93 \times 10^3$            & $4.96 \times 10^2$      \\
California Housing  & $9.68 \times 10^5$   & $2.85 \times 10^6$     & $1.75 \times 10^6$            & $8.76 \times 10^4$      \\
Delta Ailerons     & $4.33 \times 10^4$    & $2.01 \times 10^5$     & $2.59 \times 10^5$            & $1.30 \times 10^4$      \\
LA Ozone           & $1.95 \times 10^1$    & $6.21 \times 10^1$     & $7.07 \times 10^3$            & $3.54 \times 10^2$      \\
Machine CPU        & $6.74 \times 10^0$    & $4.79 \times 10^1$     & $2.22 \times 10^3$            & $1.11 \times 10^2$      \\
Prostate Cancer    & $3.29 \times 10^0$    & $1.30 \times 10^1$     & $2.34 \times 10^2$            & $1.17 \times 10^1$      \\
Servo              & $6.24 \times 10^0$    & $8.03 \times 10^1$     & $1.16 \times 10^3$            & $5.82 \times 10^1$      \\ 
\botrule
\end{tabular}
\end{table}

\section{Conclusion}\label{sec: conclusion}

In this work, we introduced a novel method for non-parametric regression tasks called the Effective Non-Random Extreme Learning Machine (ENR-ELM), designed to address key issues present in traditional Extreme Learning Machines (ELM). Our approach selects the hidden layer weights in a data-dependent manner, which facilitates model selection, overcomes challenges associated with the random weight initialization of conventional ELMs, and eliminates the need for computationally expensive inverse matrix calculations to determine the output layer weights.

Our research followed a recent trend in literature involving  a data-dependent hidden layer transformation. Specifically, our approach aims to generate the hidden layer features such that they approximate an orthonormal basis for the feature space generated by an infinite number of hidden layer neurons. We derive this target basis using the eigen-decomposition of the Gram matrix of such infinitely wide hidden layer transformation.

With this data-dependent hidden layer transformation in place, we proposed two distinct approaches for utilizing the generated features. The ``approximated" approach operates under the assumption that the transformed hidden layer features perfectly match the desired orthonormal basis. Although this assumption is not strictly accurate, the method remains effective, thus the ``approximated" designation. In contrast, the ``incremental" approach relies solely on the actual features produced by the hidden layer transformation. Since the resulting design matrix for the Ordinary Least Squares problem in the output layer may not be full rank, we apply an iterative algorithm that adaptively weights each feature’s contribution.

Our empirical evaluation demonstrated that the proposed ENR-ELM significantly reduces the overall computational time required for both model selection and training, without sacrificing predictive performance. By introducing this data-dependent method for selecting hidden layer weights, we advance the field of neural network architectures that do not depend on backpropagation. While backpropagation remains the standard for achieving highly effective results, it is also computationally intensive. Our method provides an efficient alternative, highlighting the potential of non-backpropagation-based approaches.

Extreme learning itself is not necessarily limited to the 1-layer architecture; it can be extended to any type of MLP (MultiLayer Perceptron). In any case, the only weights learned from the data are those of the output layer, while all the weights of the hidden layers are left untrained. In this way, the model in eq.~\eqref{eq: regression_function} for the regression function remains the same, the random projection $z(\bold x)$ defined in eq.~\eqref{eq: random_proj} are instead obtained as a concatenation of all the hidden layers. Since the addition of intermediate layers does not entail any difference in the mathematical treatment, we decided to present only ELM for simplicity of presentation. Finally, though ELM unifies classification and regression tasks, we only focused on regression. However we do not see particular difficulty in extending our proposed method to the classification task, apart from substituting the $l_2$ loss in eq.~\eqref{eq: betahat_def} with a class separability loss.

\section*{Declarations}


\begin{itemize}
\item Funding: Daniela De Canditiis was partially funded by "INdAM - GNCS Project", codice CUP-E53C23001670001
\item The authors declare no conflict of interest
\item Code availability: the code is available at https://github.com/FabianoVeglianti/Effective-Non-Random-
Extreme-Learning-Machine. 
\end{itemize}

\noindent

\bigskip
\begin{flushleft}%

\bigskip\noindent

\bigskip\noindent

\bigskip\noindent

\bigskip\noindent
\end{flushleft}

\newpage

\begin{appendices}

\section{Additional Results}\label{secA1}

This section provides additional plots and tables for the remaining synthetic data-sets.

\begin{figure}[H]
\centering
\includegraphics[width=\textwidth,height=\textheight,keepaspectratio]{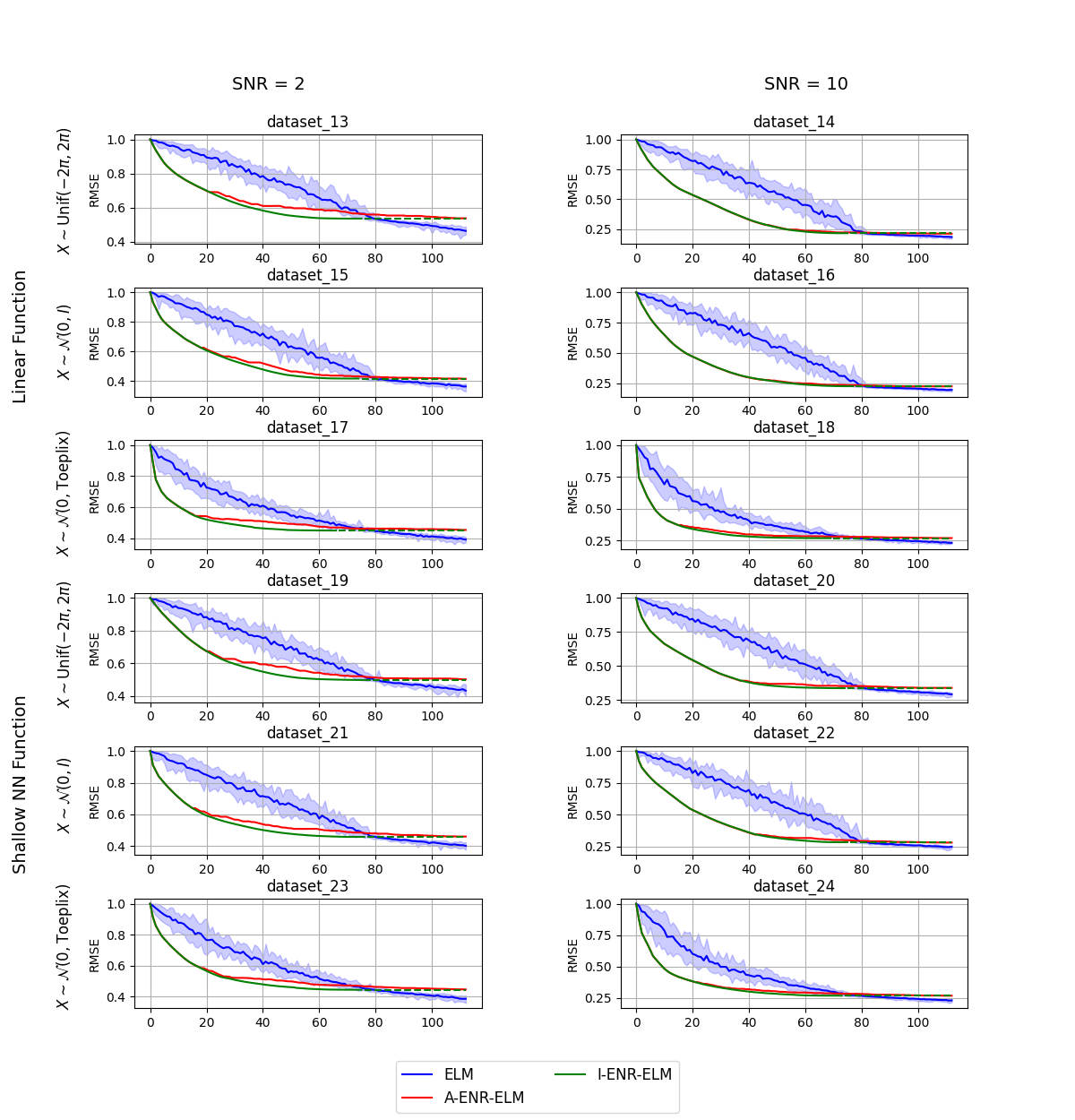}
\caption{Training error of ELM, incremental-ENR-ELM, and approximated-ENR-ELM on synthetic datasets with sample size $T = 300$ and input dimension $n_0=80$. The ELM curve represents the mean training error over 20 independent realizations of the ELM model, with the shaded region indicating the range between the minimum and maximum values observed across these realizations.}\label{fig: synthetic_datasets_training_300_80}
\end{figure}

\begin{figure}[H]
\centering
\includegraphics[width=\textwidth,height=\textheight,keepaspectratio]{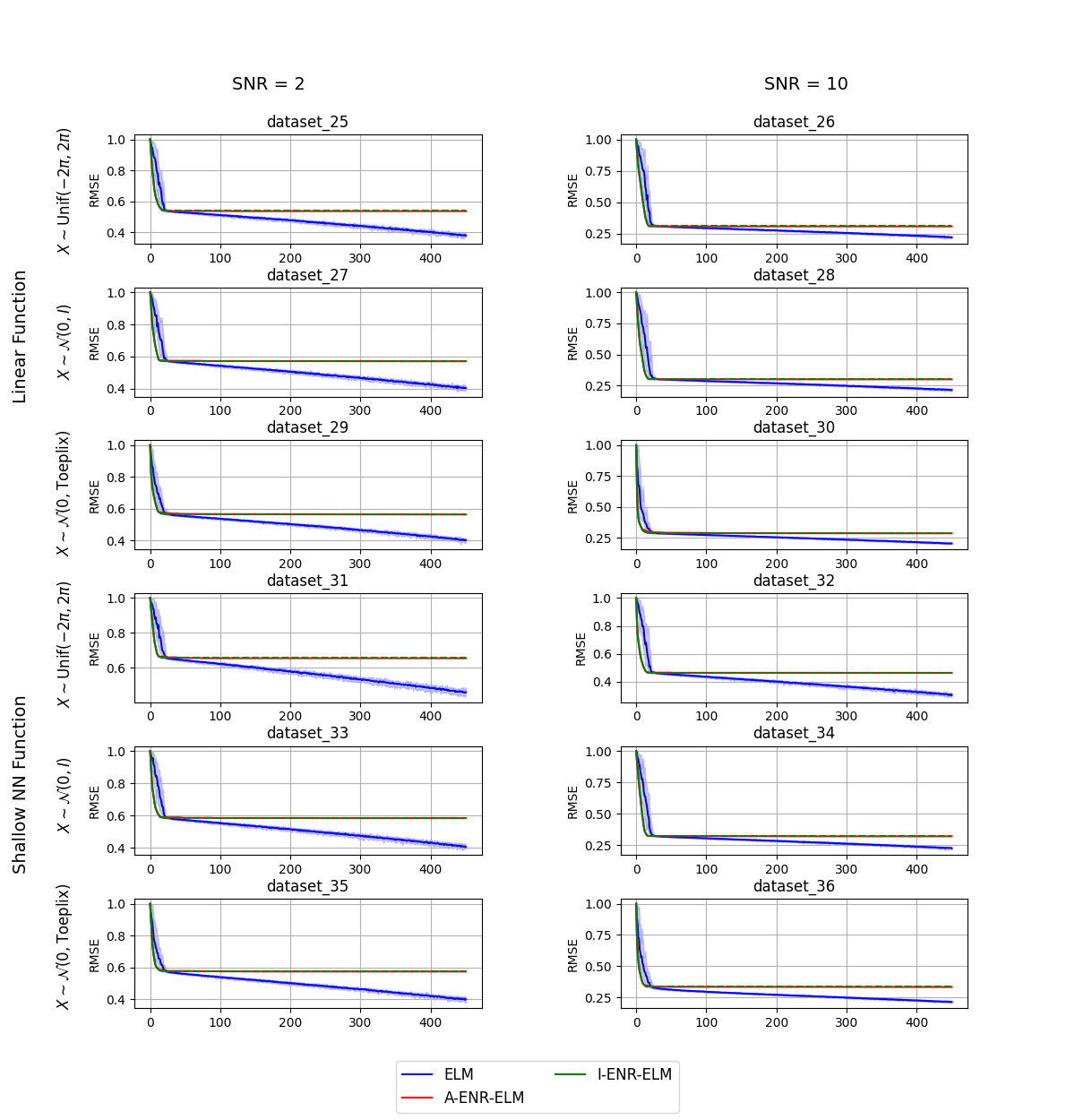}
\caption{Training error of ELM, incremental-ENR-ELM, and approximated-ENR-ELM on synthetic datasets with sample size $T = 1200$ and input dimension $n_0=20$. The ELM curve represents the mean training error over 20 independent realizations of the ELM model, with the shaded region indicating the range between the minimum and maximum values observed across these realizations.}\label{fig: synthetic_datasets_training_1200_20}
\end{figure}

\begin{figure}[H]
\centering
\includegraphics[width=\textwidth,height=\textheight,keepaspectratio]{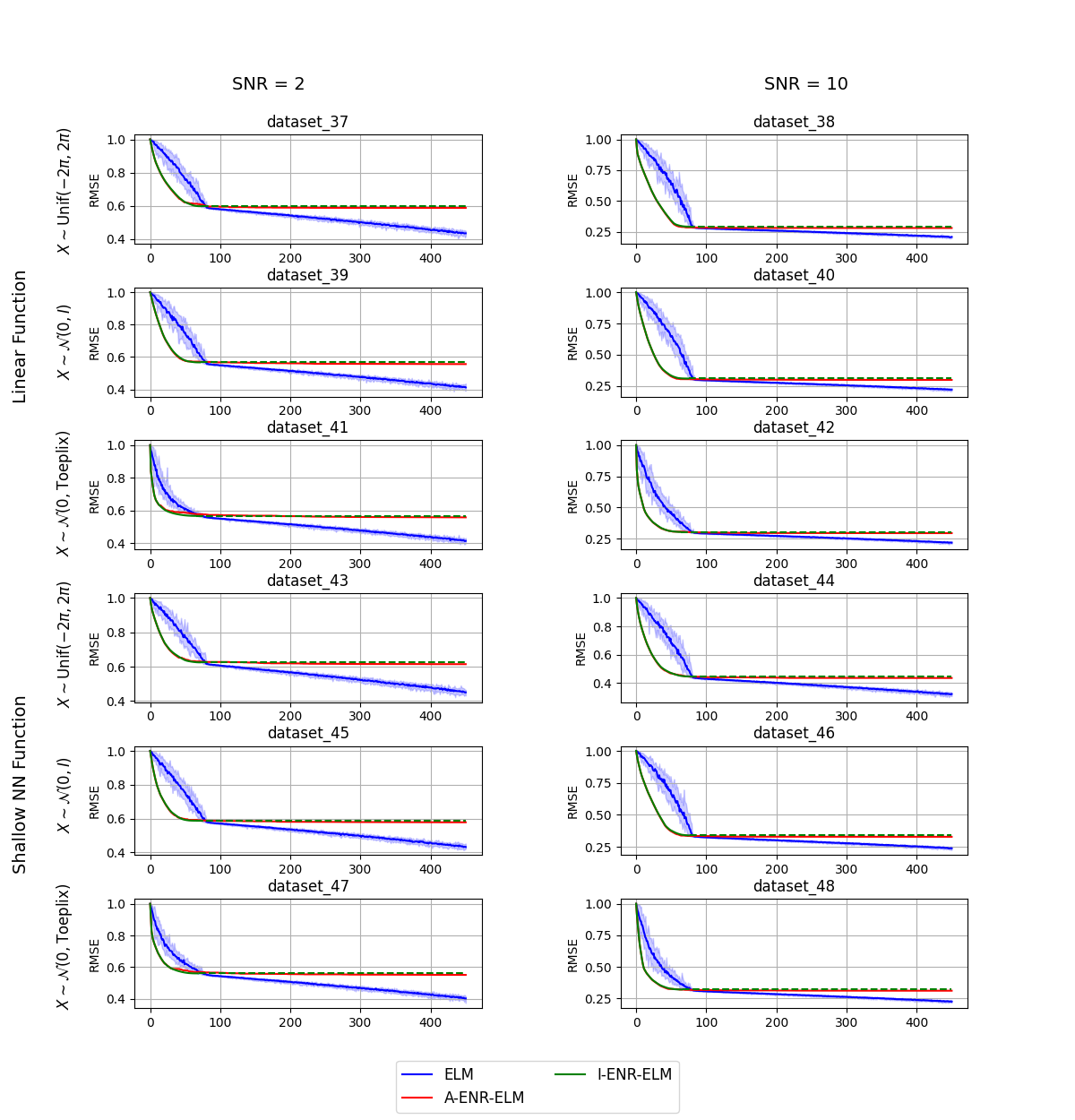}
\caption{Training error of ELM, incremental-ENR-ELM, and approximated-ENR-ELM on synthetic datasets with with sample size $T = 1200$ and input dimension $n_0=80$. The ELM curve represents the mean training error over 20 independent realizations of the ELM model, with the shaded region indicating the range between the minimum and maximum values observed across these realizations.}\label{fig: synthetic_datasets_training_1200_80}
\end{figure}

\begin{figure}[H]
\centering
\includegraphics[width=\textwidth,height=\textheight,keepaspectratio]{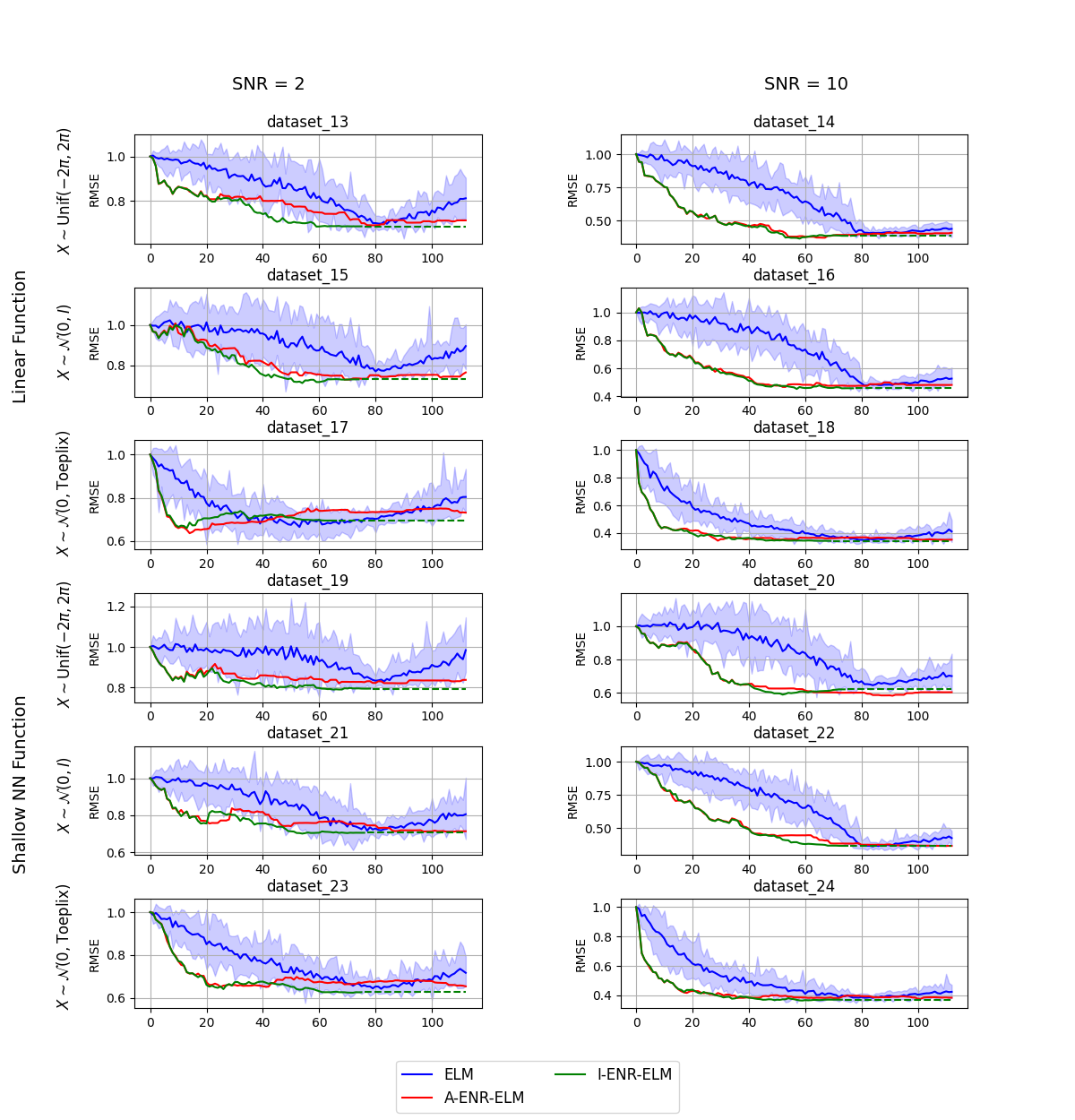}
\caption{Test error of ELM, incremental-ENR-ELM, and approximated-ENR-ELM on synthetic datasets with with sample size $T = 300$ input dimension of $n_0=80$. The ELM curve represents the mean test error over 20 independent realizations of the ELM model, with the shaded region indicating the range between the minimum and maximum values observed across these realizations.}\label{fig: synthetic_datasets_test_300_80}
\end{figure}

\begin{figure}[H]
\centering
\includegraphics[width=\textwidth,height=\textheight,keepaspectratio]{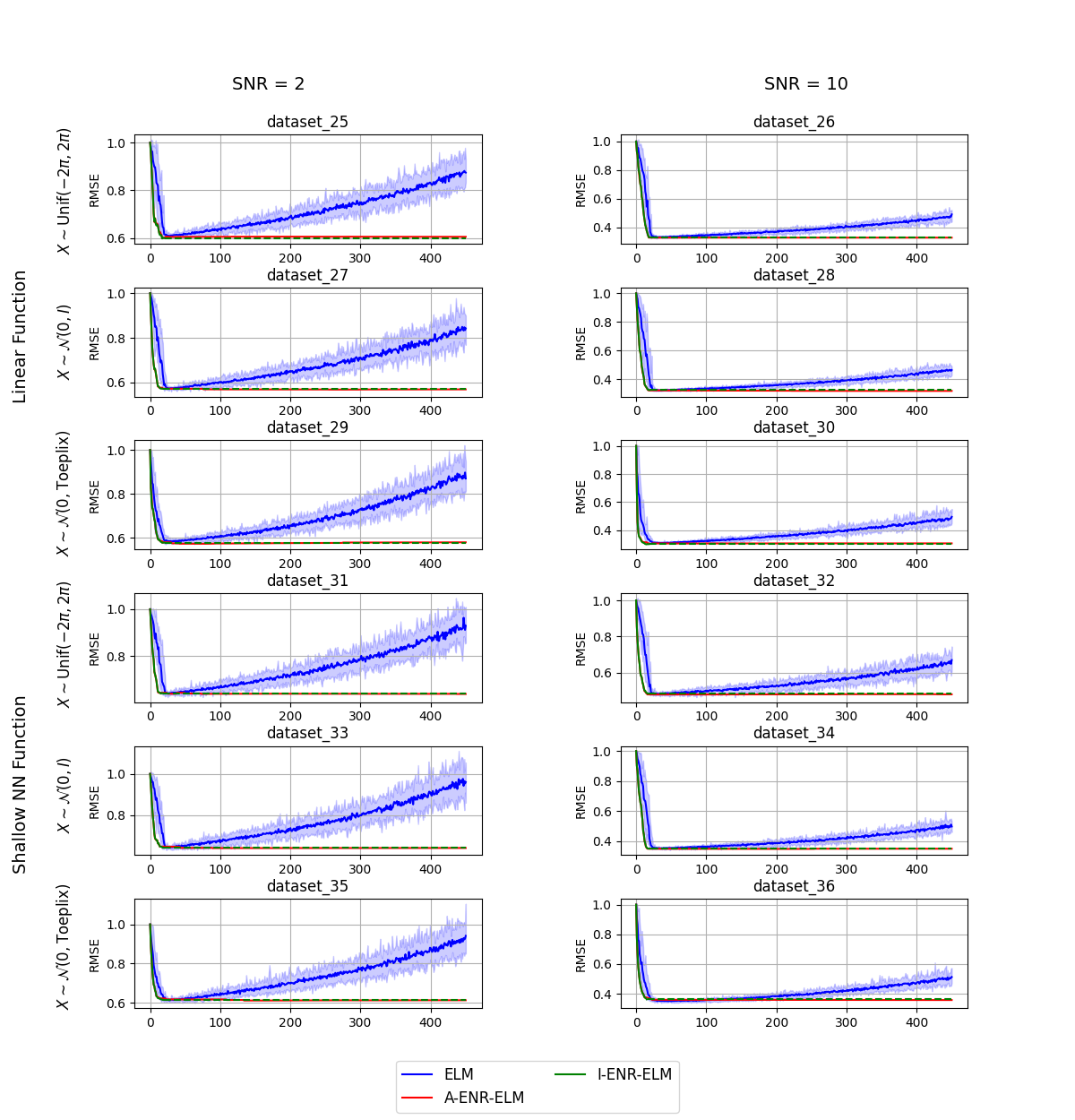}
\caption{Test error of ELM, incremental-ENR-ELM, and approximated-ENR-ELM on synthetic datasets with sample size $T = 1200$ input dimension of $n_0=20$. The ELM curve represents the mean test error over 20 independent realizations of the ELM model, with the shaded region indicating the range between the minimum and maximum values observed across these realizations.}\label{fig: synthetic_datasets_test_1200_20}
\end{figure}

\begin{figure}[H]
\centering
\includegraphics[width=\textwidth,height=\textheight,keepaspectratio]{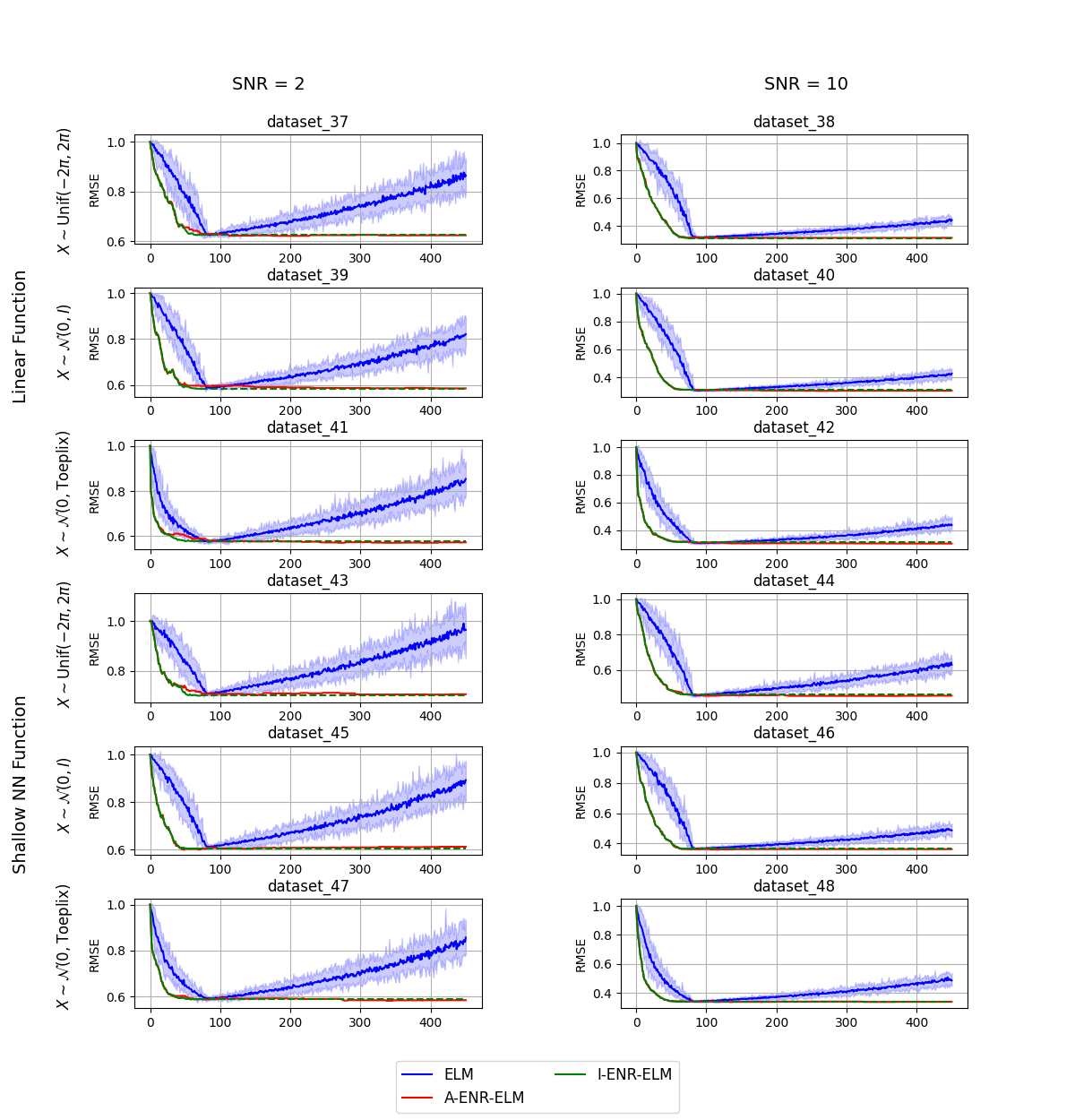}
\caption{Test error of ELM, incremental-ENR-ELM, and approximated-ENR-ELM on synthetic datasets with with sample size $T = 1200$ input dimension of $n_0=80$. The ELM curve represents the mean test error over 20 independent realizations of the ELM model, with the shaded region indicating the range between the minimum and maximum values observed across these realizations.}\label{fig: synthetic_datasets_test_1200_80}
\end{figure}

\newpage

\begin{table}[h!]
\caption{Optimal number of hidden layer neurons and relative test error (the minimum of the test error curve) for synthetic datasets with sample size $T=300$ and input dimension $n_0=80$. For the traditional ELM, the minimum is evaluated on the mean test error curve across 20 realizations and its standard deviation is reported.}
\label{table: performance-synthetic-13-24}
\begin{tabular}{l|cc|cc|cc}
\toprule
\multirow{2}{*}{\textbf{dataset}} & \multicolumn{2}{c|}{\textbf{A-ENR-ELM}} & \multicolumn{2}{c|}{\textbf{I-ENR-ELM}} & \multicolumn{2}{c}{\textbf{ELM}} \\
 & \textbf{n} & \textbf{value} & \textbf{n} & \textbf{value} & \textbf{n} & \textbf{value} \\
\midrule
Dataset 13 & 78  & 0.6902  & 59  & 0.6844  & 83  & 0.6969 $\pm$ 0.0136 \\
Dataset 14 & 66  & 0.3728  & 58  & 0.3659  & 90  & 0.4068 $\pm$ 0.0118 \\
Dataset 15 & 68  & 0.7294  & 54  & 0.7125  & 79  & 0.7704 $\pm$ 0.0164 \\
Dataset 16 & 52  & 0.4669  & 57  & 0.4527  & 80  & 0.4778 $\pm$ 0.0123 \\
Dataset 17 & 13  & 0.6360  & 13  & 0.6594  & 52  & 0.6649 $\pm$ 0.0273 \\
Dataset 18 & 28  & 0.3448  & 69  & 0.3433  & 80  & 0.3491 $\pm$ 0.0123 \\
Dataset 19 & 87  & 0.8214  & 65  & 0.7903  & 82  & 0.8306 $\pm$ 0.0214 \\
Dataset 20 & 90  & 0.5840  & 52  & 0.5906  & 83  & 0.6466 $\pm$ 0.0114 \\
Dataset 21 & 106 & 0.7112  & 54  & 0.7005  & 76  & 0.7160 $\pm$ 0.0303 \\
Dataset 22 & 111 & 0.3656  & 75  & 0.3661  & 83  & 0.3619 $\pm$ 0.0130 \\
Dataset 23 & 24  & 0.6508  & 72  & 0.6238  & 81  & 0.6419 $\pm$ 0.0131 \\
Dataset 24 & 99  & 0.3770  & 59  & 0.3649  & 83  & 0.3829 $\pm$ 0.0111 \\
\bottomrule
\end{tabular}
\end{table}

\begin{table}[h!]
\caption{Optimal number of hidden layer neurons and relative test error (the minimum of the test error curve) for synthetic datasets with sample size $T=1200$ and input dimension $n_0=20$. For the traditional ELM the minimum is evaluated on the mean test error curve across 20 realizations and its standard deviation is reported.}
\label{table: performance-synthetic-25-36}
\begin{tabular}{l|cc|cc|cc}
\toprule
\multirow{2}{*}{\textbf{dataset}} & \multicolumn{2}{c|}{\textbf{A-ENR-ELM}} & \multicolumn{2}{c|}{\textbf{I-ENR-ELM}} & \multicolumn{2}{c}{\textbf{ELM}} \\
 & \textbf{n} & \textbf{value} & \textbf{n} & \textbf{value} & \textbf{n} & \textbf{value} \\
\midrule
Dataset 25  & 20   & 0.6003  & 17   & 0.5977  & 27   & 0.6083 $\pm$ 0.0032 \\
Dataset 26  & 247  & 0.3276  & 20   & 0.3304  & 30   & 0.3307 $\pm$ 0.0031 \\
Dataset 27  & 269  & 0.5687  & 16   & 0.5729  & 25   & 0.5727 $\pm$ 0.0032 \\
Dataset 28  & 397  & 0.3171  & 19   & 0.3233  & 31   & 0.3207 $\pm$ 0.0025 \\
Dataset 29  & 78   & 0.5731  & 17   & 0.5776  & 30   & 0.5839 $\pm$ 0.0048 \\
Dataset 30  & 21   & 0.3038  & 15   & 0.3001  & 34   & 0.3108 $\pm$ 0.0026 \\
Dataset 31  & 439  & 0.6399  & 15   & 0.6415  & 26   & 0.6422 $\pm$ 0.0042 \\
Dataset 32  & 40   & 0.4796  & 17   & 0.4842  & 26   & 0.4829 $\pm$ 0.0035 \\
Dataset 33  & 233  & 0.6389  & 18   & 0.6412  & 25   & 0.6412 $\pm$ 0.0030 \\
Dataset 34  & 157  & 0.3478  & 19   & 0.3502  & 33   & 0.3510 $\pm$ 0.0029 \\
Dataset 35  & 138  & 0.6113  & 18   & 0.6150  & 26   & 0.6128 $\pm$ 0.0043 \\
Dataset 36  & 362  & 0.3579  & 15   & 0.3630  & 49   & 0.3498 $\pm$ 0.0047 \\
\bottomrule
\end{tabular}
\end{table}

\newpage

\begin{table}[h!]
\caption{Optimal number of hidden layer neurons and relative test error (the minimum of the test error curve) for synthetic datasets with sample size $T=1200$ and input dimension $n_0=80$. For the traditional ELM the minimum is evaluated on the mean test error curve across 20 realizations and its standard deviation is reported.}
\label{table: performance-synthetic-37-48}
\begin{tabular}{l|cc|cc|cc}
\toprule
\multirow{2}{*}{\textbf{dataset}} & \multicolumn{2}{c|}{\textbf{A-ENR-ELM}} & \multicolumn{2}{c|}{\textbf{I-ENR-ELM}} & \multicolumn{2}{c}{\textbf{ELM}} \\
 & \textbf{n} & \textbf{value} & \textbf{n} & \textbf{value} & \textbf{n} & \textbf{value} \\
\midrule
Dataset 37 & 202 & 0.6208 & 66 & 0.6242 & 84 & 0.6245 $\pm$ 0.0037 \\
Dataset 38 & 190 & 0.3132 & 72 & 0.3131 & 101 & 0.3163 $\pm$ 0.0029 \\
Dataset 39 & 449 & 0.5847 & 71 & 0.5830 & 85 & 0.5862 $\pm$ 0.0029 \\
Dataset 40 & 265 & 0.3020 & 70 & 0.3108 & 87 & 0.3048 $\pm$ 0.0021 \\
Dataset 41 & 389 & 0.5705 & 60 & 0.5782 & 82 & 0.5751 $\pm$ 0.0037 \\
Dataset 42 & 420 & 0.3034 & 59 & 0.3144 & 91 & 0.3063 $\pm$ 0.0026 \\
Dataset 43 & 152 & 0.7011 & 69 & 0.6997 & 86 & 0.7078 $\pm$ 0.0053 \\
Dataset 44 & 318 & 0.4545 & 73 & 0.4623 & 83 & 0.4563 $\pm$ 0.0041 \\
Dataset 45 & 48 & 0.6008 & 50 & 0.6047 & 84 & 0.6115 $\pm$ 0.0042 \\
Dataset 46 & 106 & 0.3603 & 69 & 0.3656 & 88 & 0.3656 $\pm$ 0.0029 \\
Dataset 47 & 336 & 0.5816 & 58 & 0.5873 & 82 & 0.5893 $\pm$ 0.0034 \\
Dataset 48 & 133 & 0.3376 & 58 & 0.3392 & 82 & 0.3406 $\pm$ 0.0027 \\
\bottomrule
\end{tabular}
\end{table}

\newpage




\end{appendices}


\bibliography{sn-bibliography}

\end{document}